\def\debug{1}
\documentclass{article} 
\usepackage{iclr2026_conference,times}

\usepackage{amsmath,amsfonts,bm}

\def\eqref#1{equation~\ref{#1}}

\def\1{\bm{1}}

\DeclareMathAlphabet{\mathsfit}{\encodingdefault}{\sfdefault}{m}{sl}
\SetMathAlphabet{\mathsfit}{bold}{\encodingdefault}{\sfdefault}{bx}{n}

\usepackage[colorlinks=true,linkcolor=black,citecolor={blue!50!black},urlcolor=black]{hyperref}
\usepackage{cleveref}
\usepackage{url}
\usepackage{graphicx}
\usepackage{enumitem}
\usepackage[table,xcdraw]{xcolor}
\usepackage{multirow}
\usepackage{tabularx}
\usepackage{booktabs}
\usepackage{array} 
\usepackage[normalem]{ulem}

\newcommand{\Cnotes}[1]{\ifnum\debug=1{\color{red} [CN: #1]}\fi}

\newcommand{\Exp}{\mathop{\mathbb{E}}}

\newcommand{\cD}{\mathcal{D}}
\newcommand{\cM}{\mathcal{M}}
\newcommand{\cN}{\mathcal{N}}

\newcommand{\bS}{\mathbf{S}}

\newcommand{\bc}{\mathbf{c}}
\newcommand{\bs}{\mathbf{s}}
\newcommand{\bt}{\mathbf{t}}
\newcommand{\bu}{\mathbf{u}}

\newcommand{\by}{\mathbf{y}}
\newcommand{\bz}{\mathbf{z}}

\newcommand{\Real}{\mathbb{R}}
\newcommand{\ID}{\textsf{ID}}
\newcommand{\proj}{\textsf{proj}}
\newcommand{\conv}{\textsf{Hull}}
\newcommand{\eff}{{\textsf{eff}}}
\newcommand{\diag}{{\textsf{diag}}}
\newcommand{\crit}{\textsf{crit}}

\usepackage[many]{tcolorbox} 
\tcbset{
    sharp corners,
    colback = white,
    before skip = 0.2cm,    
    after skip = 0.5cm      
}                           
\newtcolorbox{boxA}{
    boxrule = 1.5pt,
    colframe = black 
}

\title{Diagnosing Generalization Failures from\\Representational Geometry Markers}

\author{Chi-Ning Chou\\
Flatiron Institute\\
\And
Artem Kirsanov \\
Harvard University\\
\And
Yao-Yuan Yang \\
Google DeepMind \\
\And
SueYeon Chung\\
Harvard University
}

\iclrfinalcopy 
\begin{document}

\maketitle
\renewcommand{\thefootnote}{\alph{footnote}}
\footnotetext{Contact: \texttt{\{cchou,schung\}@flatironinstitute.org}}
\renewcommand{\thefootnote}{\arabic{footnote}}

\begin{abstract}
Generalization—the ability to perform well beyond the training context—is a hallmark of biological and artificial intelligence, yet anticipating unseen failures remains a central challenge. Conventional approaches often take a ``bottom-up'' mechanistic route by reverse-engineering interpretable features or circuits to build explanatory models. 
While insightful, these methods often struggle to provide the high-level, predictive signals for anticipating failure in real-world deployment. 
Here, we propose using a ``top-down'' approach to studying generalization failures inspired by medical biomarkers: identifying system-level measurements that serve as robust indicators of a model’s future performance. 
Rather than mapping out detailed internal mechanisms, we systematically design and test network markers to probe structure–function links, identify prognostic indicators, and validate predictions in real-world settings. 
In image classification, we find that task-relevant geometric properties of in-distribution (ID) object manifolds consistently forecast poor out-of-distribution (OOD) generalization. In particular, reductions in two geometric measures—effective manifold dimensionality and utility—predict weaker OOD performance across diverse architectures, optimizers, and datasets. We apply this finding to transfer learning with ImageNet-pretrained models. We consistently find that the same geometric patterns predict OOD transfer performance more reliably than ID accuracy. This work demonstrates that representational geometry can expose hidden vulnerabilities, offering more robust guidance for model selection and AI interpretability.

\end{abstract}

\section{Introduction}
Biomarkers—like blood pressure or cholesterol levels—are indispensable tools for anticipating health risks before symptoms emerge.
Throughout the history of medicine, physicians have often utilized these diagnostic measures effectively before figuring out all the biological details.~\footnote{The lipid hypothesis, for instance, linked cholesterol to cardiovascular disease risk well before lipid pathways were mapped. Similarly, selective serotonin reuptake inhibitors treated depression, informed by the serotonin hypothesis, decades before serotonin's precise role in mood regulation was fully understood.} 
This pragmatic, top-down approach of correlating biomarkers with outcomes has thus driven medical progress, while simultaneously providing the foundational insights for figuring out causal mechanisms.
In neuroscience, the same methodology has been fruitful: single-neuron and population-level signatures have served as useful analysis units, revealing principles of coding and computation often before a full mechanistic understanding~\citep{rigotti2013importance,barak2013sparseness,mastrogiuseppe2018linking,stringer2019high}.

As deep neural networks (DNNs) become increasingly integrated into critical applications, a similar challenge arises: how can we anticipate their unseen failures? This is particularly important under distribution shifts where training and deployment environments differ~\citep{sagawa2020distributionallyrobustneuralnetworks,liu2021towards,yang2024generalized}. Current research often favors bottom-up approaches, such as mechanistic interpretability (MI), which aims to reverse-engineer DNNs by identifying interpretable features \citep{olah2017feature,yun2023transformervisualizationdictionarylearning,cunningham2023sparseautoencodershighlyinterpretable}, functional circuits \citep{Olah2020ZoomIA,dunefsky2024transcoders}, or causal structures \citep{mueller2024quest,geiger2025causal}. 
While these MI methods offer granular insights, they may lack identifiability \citep{meloux2025everything}, and it remains unclear how they can provide concrete diagnostics on real-world models. 

\begin{figure}[ht!]
    \centering
    \includegraphics[width=\linewidth]{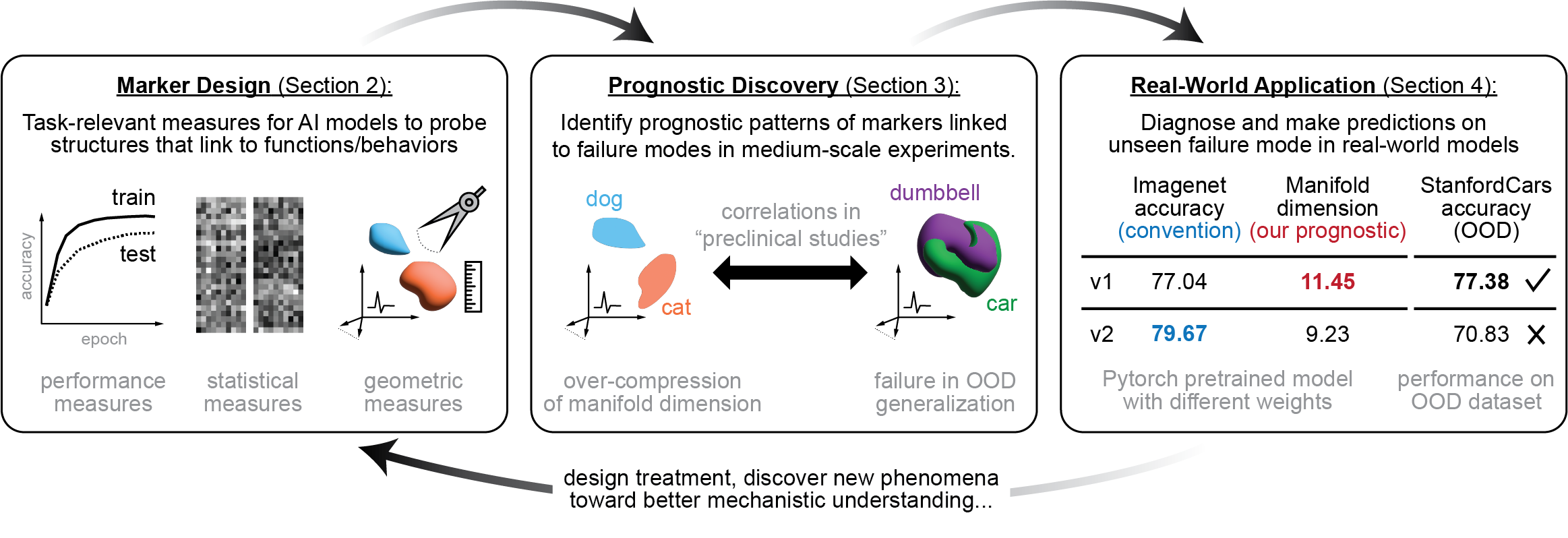}
    \vspace{-3mm}
    \caption{A diagnostic, system-level paradigm for studying generalization failures in DNNs, with an example on image classification. See~\autoref{sec:intro overview} for an overview.
    }
    \label{fig:paradigm}
\end{figure}

Here we propose a complementary perspective inspired by the history of medicine: a diagnostic, system-level paradigm for understanding neural networks. Rather than attempting to reconstruct their internal mechanisms post-hoc, we focus on developing task-relevant measurements—markers for AI models—that serve as reliable indicators of potential failure modes. Our methodology follows a three-step cycle (\autoref{fig:paradigm}):
(i) \textbf{Marker Design}: develop task-relevant measures to probe which structures in neural networks (e.g., feature vectors, weights) relate to their function and performance;
(ii) \textbf{Prognostic Discovery}\footnote{In medicine, diagnostics identify present conditions, while prognostics forecast future risks. Our framework is termed ``diagnostic'' broadly, with Step 2 specified as ``prognostic discovery'' to emphasize prediction of OOD failures from ID data.}: conduct medium-size experiments across diverse architectures and hyperparameters, and identify patterns that serve as prognostic indicators—signals present in in-distribution (ID) properties that can forecast future generalization failures without requiring any knowledge of the out-of-distribution (OOD) tasks;
(iii) \textbf{Real-world Application}: apply these insights to practical settings, such as predicting which pretrained models will transfer more robustly across datasets.
We demonstrate this research cycle by using ID measures based on task-relevant representational geometry to diagnose failure in OOD generalization. Our framework points toward a diagnostic science of AI models, offering tools to anticipate vulnerabilities and improve robustness in safety-critical domains.

\vspace{-3mm}
\subsection{Overview and our contributions}\label{sec:intro overview}
In this work, we apply the proposed diagnostic, system-level paradigm to investigate failure modes of OOD generalization in image classification. Our key finding is that feature overspecialization—quantified by reduced effective dimensionality and utility of object manifolds—is a reliable indicator of poor performance under class-level distribution shifts in transfer learning.

\vspace{-3mm}
\paragraph{Marker design for image classification (\autoref{sec:biomarker}).}
A central step in our diagnostic framework is selecting and designing \textit{markers}—scalar quantities computed entirely from ID data—that capture aspects of a pretrained model relevant for downstream generalization.
In image classification, we focus on penultimate-layer feature vectors, as the final classification decision is obtained by a linear readout from this layer.
Accordingly, we evaluate a broad family of candidate markers, including:
(i) accuracy- and logits-based quantities,
(ii) low-order statistical summaries of representations (e.g., sparsity, covariance structure), and
(iii) geometric measures of class-conditioned point-cloud manifolds—such as participation ratio, within-class spread, neural-collapse metrics~\citep{papyan2020prevalence,haruncontrolling}, numerical rank~\citep{masarczyk2023tunnel,harun2024variables}, and task-relevant geometric measures from the GLUE framework~\citep{chou2025glue}.

\paragraph{Prognostic discovery of OOD generalization failures (\autoref{sec:prognostic}).}
We conducted exploratory experiments to investigate whether these metrics can predict failures in OOD generalization. Specifically, we trained a broad class of deep networks on in-distribution (ID) data (e.g., CIFAR-10) and evaluated their OOD performance on datasets with disjoint classes (e.g., CIFAR-100; Figure~\ref{fig:main schematic}a,b). Our sweep spanned five architectures (e.g., ResNet, VGG), multiple depths, two optimization algorithms (SGD, AdamW), and a grid of hyperparameters (learning rate, weight decay). 
We found that different training hyperparameters can lead to markedly different OOD performance, despite nearly identical ID train and test accuracy. Task-relevant geometric measures of ID object manifolds correlated far more strongly with OOD performance than conventional performance metrics (e.g., ID accuracy) or statistical measures (e.g., sparsity, covariance) (\autoref{fig:main schematic}c, top). In particular, reductions in effective dimensionality and utility consistently served as prognostic indicators of OOD failure (\autoref{fig:main schematic}c, bottom). 
Together with prior work linking representational geometry to feature learning~\citep{chou2025featurelearninglazyrichdichotomy}, these findings suggest that overspecialized features undermine generalization performance, echoing previous accounts of shortcut learning~\citep{geirhos2020shortcut}.

\vspace{-3mm}
\paragraph{Applications to failure prediction in pretrained models (\autoref{sec:application}).}
Finally, we applied our prognostic indicators to ImageNet-pretrained models from public repositories. In practice, when selecting among multiple pretrained weights of the same architecture, the most common criterion is test accuracy. Here, we measured the effective dimensionality and utility of ImageNet object manifolds from 20 architectures available in PyTorch (e.g., RegNet, MobileNet, WideResNet), each released with two weights (v1 and v2); by construction, v2 achieves higher ID accuracy. Unlike our controlled prognostic studies, these pretrained weights were produced under distinct training recipes, regularization schemes, and preprocessing pipelines, making them a much more heterogeneous testbed. 
Nevertheless, consistent with the predictions from our medium-scale experiments, models where v1 exhibited higher manifold dimension and utility than v2 also achieved better OOD performance under v1 weights—even though v1 had lower ID accuracy (\autoref{tab:pretrained}). This demonstrates that ID representational geometry can serve as an early diagnostic for OOD robustness.

\textit{Summary.} Our work demonstrates a diagnostic, system-level paradigm that complements conventional mechanistic interpretability by focusing on predictive indicators of model failure. Our results highlight how task-relevant geometric measures of ID representations can serve as markers for diagnosing failures in OOD generalization, even when mechanistic details remain opaque.

\subsection{Related work}
\paragraph{Representational geometry and generalization.}
A growing body of work suggests that properties of internal representations in DNNs can indicate generalization performance. In the standard ID setting, both statistical features of activations—such as sparsity, covariance, and inter-feature correlations~\citep{morcos2018importance}—and geometric measures of object manifolds~\citep{ansuini2019intrinsic,cohen2020separability,chou2025featurelearninglazyrichdichotomy} have been predictive. For example, networks that generalize well often exhibit low intrinsic dimensionality in their final-layer representations, and such compactness correlates with test accuracy in image classification~\citep{ansuini2019intrinsic}. A related phenomenon is \textit{neural collapse}~\citep{papyan2020prevalence}, where within-class variability of final hidden representations vanishes in the terminal phase of training.

The picture becomes more convoluted under distribution shifts. \citep{galanti2022on} showed that neural collapse can generalize to new data points and classes when trained on sufficiently many classes with lots of samples. By contrast,~\citep{zhu2023variance} found that encouraging diversity and decorrelation among features improves OOD performance in image and video classification. Similarly, in neuroscience, high-dimensional yet smooth population codes in mouse visual cortex have been linked to generalization across stimulus conditions~\citep{rigotti2013importance,stringer2019high}. These conflicting results call for more systematic study on how representational properties connect to OOD generalization, and our findings—that OOD failures correlate with overcompression of object manifolds—support these results on the advantage of high-dimensional representations. However, most of these approaches rely on generic geometric or statistical descriptors that are not explicitly tied to the computational task, whereas the GLUE measures we employ use the anchor point distribution (\autoref{fig:schematic geometry}c and~\autoref{app:geom_measures}) to directly link representational geometry to downstream linear classification performance.

\paragraph{Distribution shift in ML: detection, transfer, and prior approaches.}
OOD detection methods aim to distinguish OOD samples from ID samples by examining differences in their feature representations or logit distributions. These approaches typically operate under label-preserving distribution shifts, where the input distribution changes but the class label space remains the same. In contrast, class-level OOD generalization—also called transfer learning—is substantially more challenging, since the OOD task contains entirely unseen class labels. Performance in this setting is usually assessed by training a linear probe (often on the penultimate-layer features) of a pretrained network. Several works have studied how architectural or representational factors influence this probe-based OOD performance. For example, the Tunnel Effect papers~\citep{masarczyk2023tunnel,harun2024variables} showed that the drop in OOD linear-probe accuracy across layers correlates with a drop in the numerical rank of OOD features. Similarly, Neural Collapse–based analyses~\citep{haruncontrolling} have examined how the extent of collapse (e.g., the $\mathcal{NC}1$ metric~\citep{papyan2020prevalence}) relates to OOD generalization performance across layers. 
Outside image classification, recently~\cite{li2025can} used interpretability methods to predict OOD model behavior in language tasks through ID attention patterns. 

In this work, we incorporate several of these ideas into our marker design. For OOD detection methods, many algorithms require access to OOD samples in their scoring pipelines, making them incompatible with our ID-only diagnostic setting. However, methods that rely solely on logit statistics or feature-level summaries can be adapted into scalar markers and included in our evaluation. For the Tunnel Effect and Neural Collapse lines of work, we directly implement their corresponding measures—numerical rank and feature-collapse metrics (e.g., $\mathcal{NC}1$)—and compare them against our GLUE-based markers in the prognostic analysis. It is worth noting that both the Tunnel Effect~\citep{masarczyk2023tunnel,harun2024variables} and Neural Collapse~\citep{haruncontrolling} studies primarily analyze how their measures vary across layers within the same model, rather than across different models or training configurations. Our focus, in contrast, is on comparing models trained under different hyperparameters or initialization regimes, which is the setting relevant for model selection and prognostic prediction.

\section{Markers for image classification}\label{sec:biomarker}
Given a neural network with parameters $\theta$ and an ID dataset $\cD_\ID$, we define a marker as a function that maps $(\theta,\cD_\ID)$ to a scalar value indicative of potential failure modes in OOD generalization. Train and test accuracy are examples of such markers, but they are often non-discriminative~\citep{d2022underspecification}, propelling us to open the black box of DNNs and seek measures that are both task-relevant and discriminative.

Among the many ways to peer inside a DNN, we focus on feature embeddings. Concretely, we analyze penultimate-layer feature vectors $\{\bz_i\}_{i=1}^M$ (e.g., \texttt{avgpool} in ResNet, see~\autoref{tab:layer_selection}) extracted from the ID data.
Each $\bz_i \in \Real^N$ is an $N$-dimensional feature vector, and in image classification these can be grouped by class: letting $P$ denote the number of classes and $M^\mu$ the number of samples in class $\mu$, we write $\{\bz_i^\mu\}_{i=1}^{M^\mu}$ so that $\{\bz_i\}_{i=1}^M = \bigcup_{\mu=1}^P \{\bz_i^\mu\}_{i=1}^{M^\mu}$. 
In addition to geometric markers derived from representational manifolds, we also consider conventional ID-only markers inspired by prior OOD-detection methods. These include low-order statistical summaries of penultimate representations (e.g., sparsity, covariance structure, pairwise distances and angles) as well as quantities computed directly from the logits distribution (e.g., averaged confidence). 

In the remainder of this section, we first review the conventional statistical and logits-based markers that serve as baselines in our analysis (\autoref{sec:biomark conventional}), and then introduce task-relevant geometric markers grounded in representational manifold theory (\autoref{sec:biomarker task relevant}).

\subsection{Conventional measures}\label{sec:biomark conventional}
We also examine low-order statistics of penultimate feature vectors.
We consider several standard statistics: activation sparsity, off-diagonal covariance magnitude, and mean pairwise distance/angle. Each measure is applied both globally across $\{\bz_i\}$ and within each class $\{\bz_i^\mu\}$. 
These descriptors summarize the distribution of representations but do not capture per-class manifold geometry, motivating the measures introduced next. 
Formal definitions are given in~\autoref{app:stat_measures}.

In addition to feature-level statistics, we incorporate several logit-based markers commonly used in OOD-detection research and adapt them to our ID-only diagnostic setting, including averaged confidence (AUROC)~\citep{hendrycks2018benchmarking}, Entropy~\citep{guillory2021predicting}, and
Energy~\citep{liu2020energy}.
While these methods were originally designed to detect OOD inputs—and typically assume access to shifted data—we evaluate them here as scalar markers derived solely from ID logits.

\vspace{-3mm}
Beyond these statistical and logits-based metrics, prior work has also analyzed representational geometry in neural populations~\citep{chung2021neural, li2024representations}. Whereas statistical metrics capture overall spread or pairwise correlations, geometric descriptors characterize manifold structure such as alignment, curvature, and class-specific variability. A widely used task-agnostic geometric marker is the participation ratio (PR), which estimates the intrinsic dimensionality of each class manifold from the spectrum of its covariance matrix. 
We also include Neural Collapse measures~\citep{papyan2020prevalence,ammar2023neco,haruncontrolling} and the numerical rank measure from the Tunnel Effect hypothesis~\citep{masarczyk2023tunnel,harun2024variables}.
Formal definitions for all markers are provided in~\autoref{app:measures}.

\subsection{Task-relevant geometric measures}\label{sec:biomarker task relevant}
\begin{figure}[t]
    \centering
    \includegraphics[width=\linewidth]{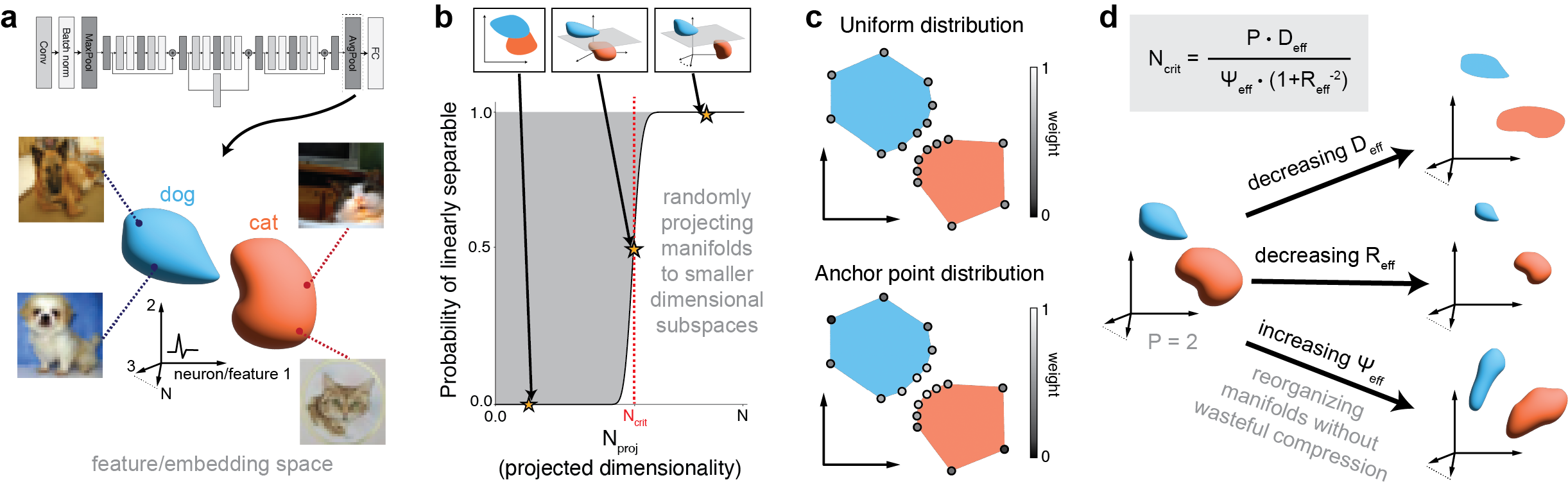}
    \caption{\textbf{Object manifolds and task-relevant geometric measures.}
    \textbf{a}, Object manifolds are the per-class point clouds in the feature space.
    \textbf{b}, Critical dimension $N_\crit$ quantifies the degree of manifold untangling/separability in an average-case sense via random projection.
    \textbf{c}, Anchor point distribution gives higher weight to points that are more important for linear classification.\protect\footnotemark \textbf{d}, The degree of manifold separation (quantified by critical number of neurons $N_\crit$) is analytically linked to three task-relevant geometric measures: effective dimension $D_\eff$, radius $R_\eff$ and utility $\Psi_\eff$.
    }
    \vspace{-3mm}
    \label{fig:schematic geometry}
\end{figure}

To obtain task-relevant markers of ID representations, we adopt the \textit{Geometry Linked to Untangling Efficiency} (GLUE) framework~\citep{chou2025glue}, which builds on the theory of perceptron capacity for points~\citep{gardner1988optimal} and manifolds~\citep{chung2018classification,wakhloo2023linear,mignacco2025nonlinear,chou2025glue} from statistical physics.
Similar to support vector machine (SVM) theory~\citep{cortes1995support}, where the max-margin classifier can be expressed as a linear combination of support vectors, GLUE theory provides an analytic connection between the \textit{critical number of neurons} $N_\crit$ and the geometry of object manifolds (\autoref{fig:schematic geometry}a) through an \textit{anchor point distribution} over the object manifolds (\autoref{fig:schematic geometry}c).

\footnotetext[5]{This figure is a schematic illustration of the non-uniform, task-relevant anchor point distribution. The 2D depiction is only intuitive and can be misleading, analogous to how in high dimensions Gaussian mass concentrates on the sphere rather than at the origin.}

\paragraph{GLUE as an average-case analog of SVM.}
Consider classifying two object manifolds $\cM^1=\conv(\{\bz^1_1,\dots,\bz^1_M\})$ and $\cM^2=\conv(\{\bz^2_1,\dots,\bz^2_M\})$ in $\Real^N$, $N_\crit$ is defined as the minimum $N_\proj$ such that manifolds remain linearly separable with probability at least 0.5 after random projection to an $N_\proj$-dimensional subspace (\autoref{fig:schematic geometry}b). Manifold capacity $\alpha$ is defined as $P/N_\crit$, where $P$ is the number of manifolds. A lower value of $N_\crit$ (i.e., higher value of $\alpha$) means that the object manifolds are more separable on average.
The key result in GLUE theory is a closed-form formula for $N_\crit$:
\begin{equation}\label{eq:glue formula anchor}
N_\crit = \Exp_{\bt\sim\cN(0,I_N)}\left[\max_{\bs^1(\bt)\in\cM^1,\bs^2(\bt)\in\cM^2}\|\proj_{\textsf{span}(\{\bs^1(\bt),\bs^2(\bt)\})} \bt\|_2^2\right]
\end{equation}
where $\cN(0,I_N)$ is the isotropic Gaussian distribution in $\Real^N$, $\textsf{span}(\cdot)$ denotes linear span of a set, and $\proj$ denotes orthogonal projection.
\autoref{eq:glue formula anchor} naturally leads to defining anchor points as the maximizers of the inner optimization problem.
The anchor point distribution is a non-uniform measure over the manifolds and favors those points that are more important for downstream classification (\autoref{fig:schematic geometry}c). Hence, GLUE theory can be thought of as an average-case analog of SVM theory: whereas SVM assesses separability in the best-case scenario by leveraging the full feature space, GLUE evaluates separability under random projections, effectively averaging across many such subspaces, and hence is able to capture more complex, heterogeneous, and nuisance structure present in the data~\citep{chou2025glue,chou2025featurelearninglazyrichdichotomy}.

By exploiting symmetries in the equation, GLUE theory derives three effective geometric measures—effective dimension $D_\eff$, effective radius $R_\eff$, and effective utility $\Psi_\eff$—and reorganizes~\autoref{eq:glue formula anchor} into a simple expression (see~\autoref{app:geom_measures} for details and derivations):
\begin{equation}\label{eq:glue geometric formula}
N_\crit = \frac{P\cdot D_\eff}{\Psi_\eff\cdot(1+R_\eff^{-2})}
\end{equation}
where $P$ is the number of manifolds. Intuitively,~\autoref{eq:glue geometric formula} shows that $N_\crit$ decreases (i.e., manifolds become more separable/untangled) with smaller $D_\eff$, smaller $R_\eff$, and larger $\Psi_\eff$ (\autoref{fig:schematic geometry}d). 
Because the GLUE theory captures task-relevant structures in neural representations via the anchor point distribution (as opposed to the uniform distribution, i.e., equiprobable sampling of points), a recent work~\citep{chou2025featurelearninglazyrichdichotomy} has shown that $N_\crit$ and GLUE measures are much more discriminative than conventional measures (e.g., kernel-based methods, weight changes) in the study of feature learning.
GLUE also defines additional measures (e.g., center, axis, center–axis alignment) from the anchor point distribution, detailed in~\autoref{app:geom_measures} and omitted here for brevity. We provide intuitions for the three effective geometric measures in~\autoref{tab:effective geometry simplified} (see~\autoref{tab:effective geometry} for the full version).

\vspace{-5mm}
\begin{table}[h]
\centering
\small
\caption{Intuitions for GLUE measures.}
\label{tab:effective geometry simplified}
\vspace{2mm}
\renewcommand{\arraystretch}{1.3} 
\begin{tabular}{|p{2cm}|p{3.3cm}|p{3.5cm}|p{3.3cm}|}
\hline
& \multicolumn{1}{c|}{$D_\eff \geq 0$} &  \multicolumn{1}{c|}{$R_\eff \geq 0$} & \multicolumn{1}{c|}{$\Psi_\eff \in [0,1]$} \\ \hline
Geometric \ \ \ \ \ \ \ \ intuition & Quantify the task-relevant dimensionality of object manifolds. & Quantify the task-relevant spread within each manifold relative to their centers. & Quantify the amount of excessive compression of untangling manifolds. \\ \hline
Effect on linear separability & More separable when $D_\eff$ is small. & More separable when $R_\eff$ is small. & More separable when $\Psi_\eff$ is large. \\ \hline
Example &  $D_\eff$ equals the dimension of uncor.~random spheres & $R_\eff$ equals the radius of uncor.~random spheres & Collapsing manifolds to points yields $\Psi_\eff\to0$. \\ \hline
Interpretation in feature\text{\ \ \ \ \ \ \ \ } learning\footnotemark & Low $D_\eff$ indicates a smaller set of feature modes in use. &  Low $R_\eff$ indicates more similar feature usage across examples within a class. & Low $\Psi_\eff$ indicates inefficient compression of within-class variability. \\ \hline
\end{tabular}
\vspace{-3mm}
\end{table}

\paragraph{Connection to feature learning.}
We follow a top-down view of feature learning~\citep{chou2025featurelearninglazyrichdichotomy}, where \textit{features} are understood functionally through their consequences for computation (e.g., enabling linear separability) rather than as specific interpretable axes or neurons. This perspective emphasizes how representational geometry changes with feature usage without requiring explicit identification of the features themselves. Moreover, by thinking of a direction in the representation space as a feature (linear representation hypothesis~\citep{park2024linear}), the effective geometric measures offer interpretation in feature learning as listed in the table.

\section{Discover Prognostics for Failure in OOD Generalization}\label{sec:prognostic}
We study medium-scale models as a testbed for identifying prognostic indicators of failure modes. Our goal is to detect ID signals that reliably predict how a model will behave under distribution shift—without any access to OOD data. This departs from most existing OOD-detection methods, which typically rely on information from the shifted distribution. Our diagnostic analysis uses markers measured solely from ID properties to anticipate vulnerabilities before deployment.

\begin{figure}[ht!]
    \centering
    \includegraphics[width=\linewidth]{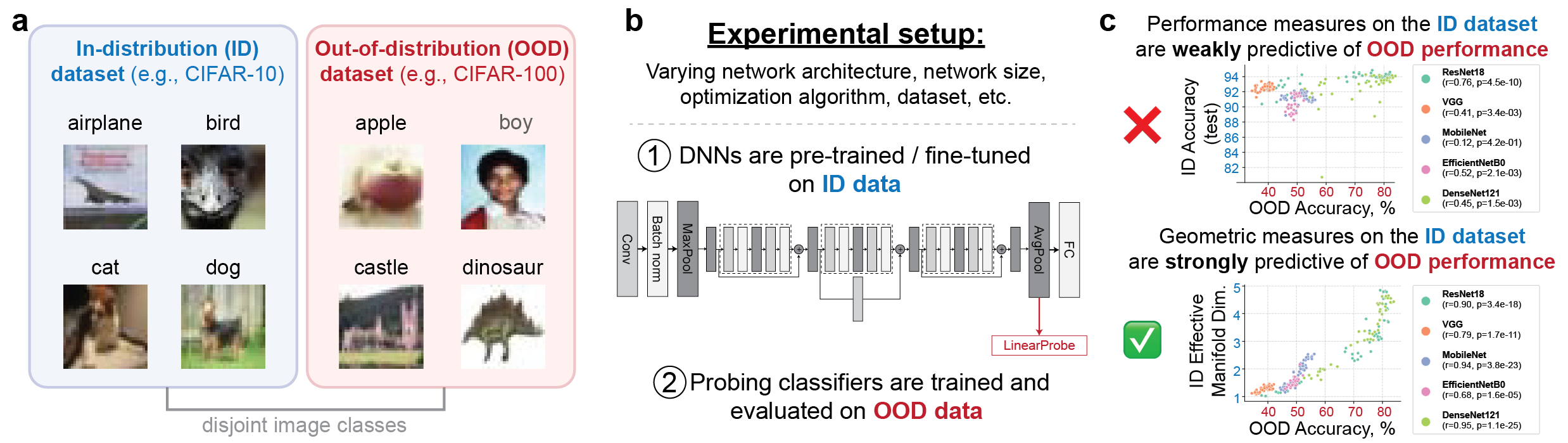}
    \caption{\textbf{Prognostic discovery for OOD generalization.}
    \textbf{a}, We consider the image classification problem with an ID dataset and an OOD dataset with disjoint image classes.
    \textbf{b}, We trained DNNs on the ID dataset and evaluated the OOD performance as linear probe accuracy.
    \textbf{c}, Conventional performance and statistical measures on the ID dataset are weakly predictive of OOD performance, while some task-relevant geometric measures can robustly predict failures in OOD generalization.
    }
    \vspace{-3mm}
    \label{fig:main schematic}
\end{figure}

\subsection{Methods}\label{sec:method}
 We adopt an experimental design in~\citep{chou2025featurelearninglazyrichdichotomy} where DNNs are trained on an ID image dataset, and OOD performance is evaluated on a different dataset with a disjoint set of classes. 

\vspace{-3mm}
\paragraph{Training procedure.}
We trained multiple DNN architectures (e.g., ResNet, VGG) from scratch on CIFAR-10. For each architecture, we swept over four initial learning rates, four weight decay values, and three random seeds, using both SGD and AdamW optimizers. In all cases, we ensured that the training accuracy was above 99\% and the test accuracy ranged from 88\% to 95\%.

\vspace{-3mm}
\paragraph{OOD evaluation via linear probing.}
To assess the OOD generalization of learned representations, we adopt a linear probing framework~\citep{alain2016understanding,zhu2023variance,chou2025featurelearninglazyrichdichotomy}.  After ID training, the network's feature extractor was frozen. A new linear classifier was then trained on top of these features using the OOD dataset. The test accuracy of this linear probe served as our measure of OOD performance (\autoref{fig:main schematic}b). See~\autoref{app:experiment} for details.

\subsection{Results}\label{sec:prognostic results}
We find that models trained with distinct hyperparameters can exhibit similar ID accuracy while their OOD performance can differ drastically. This variation, however, is not random; we find that OOD performance can be consistently predicted by geometric properties of ID representations.

\begin{figure}[ht!]
    \centering
    \includegraphics[width=\linewidth]{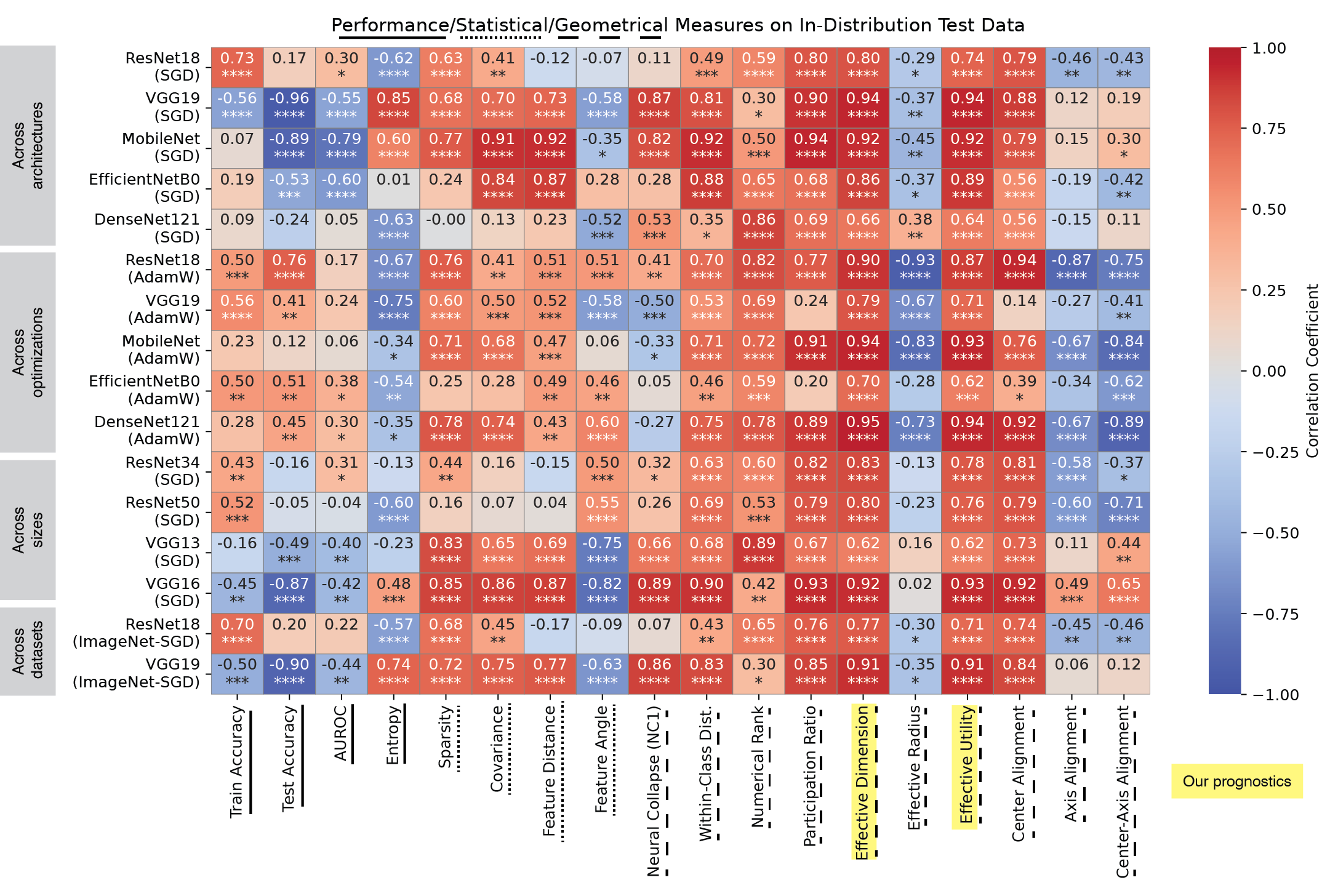}
    \vspace{-5mm}
    \caption{All results on models trained on CIFAR-10, showing correlations between markers (x-axis) and OOD performance across a hyperparameter sweep. Numbers indicate Pearson $r$; asterisks denote significance ($^{*}: p\le0.05$; $^{**}: p\le0.01$; $^{***}: p\le0.001$; $^{****}: p\le0.0001$). 
    }
    \vspace{-3mm}
    \label{fig:main across more params}
\end{figure}

\vspace{-3mm}
\paragraph{Task-relevant geometric markers are predictive across architectures.}
First, we trained different architectures (ResNet, VGG, etc) on CIFAR-10 and evaluated OOD performance on CIFAR-100. As summarized in \autoref{fig:main across more params}, conventional metrics like ID accuracy and statistical measures like sparsity showed weak and inconsistent correlations with OOD performance. In contrast, several geometric measures—particularly participation ratio, effective dimension, and effective utility—were strong predictors and consistently performed well across all architectures.

\vspace{-3mm}
\paragraph{Findings hold across model sizes, optimizers, and datasets.}
Next, we tested the generality of our findings by varying model size (ResNet18/34/50), optimizer (SGD, AdamW), and the OOD dataset (CIFAR-100, ImageNet). The results, shown in~\autoref{fig:main across more params}, remained consistent. Across all these settings, task-relevant geometric signatures of the ID data were systematically predictive of OOD performance, whereas alternative markers—including Neural Collapse~\citep{haruncontrolling}, numerical rank (Tunnel Effect~\citep{masarczyk2023tunnel}), and logits-based OOD-detection scores—showed statistically weaker or less consistently predictive trends across settings, with numerical rank performing well in most but a few cases (e.g., VGG-19 using SGD). We suspect this is because the Neural Collapse and Tunnel Effect measures were primarily designed on mathematical intuition rather than task-relevant considerations; as a result, they may not capture the fine-grained structure of complex neural activity patterns across different models or training regimes.
Conversely, logit-based markers such as AUROC or entropy are task-relevant but appear to discard too much of the rich information in internal representations, limiting their predictive power in this setting. Additional results are provided in~\autoref{app:additional_results}.

\vspace{-3mm}
\paragraph{Task-relevant geometric markers from ID training data also show strong trends.}
While the main figures report results using ID test or validation features, we find that the same geometric indicators measured directly on the ID training data exhibit similarly strong correlations with OOD performance (see~\autoref{fig:table_train}). This indicates that the predictive signal is not limited to held-out examples, but is already present in the geometry of the training representations themselves.

\vspace{-3mm}
\paragraph{ID Test accuracy best predicts OOD performance on corrupted images.}
We also consider a corrupted version (e.g., adding noise, varying brightness, pixellating, etc.) of the original images as an OOD dataset (e.g., CIFAR-10C~\citep{hendrycks2018benchmarking}).  Since the class labels remain identical to the ID dataset, OOD performance can be measured directly by the trained network, without training an additional linear probe. In this setting, ID test accuracy is the strongest predictor of performance on corrupted data (see~\autoref{app:corrupted data}), although we note that it does not always work. We also observe distinct geometric patterns across different corruption types. These results highlight that the correlation between OOD accuracy and manifold compression (\autoref{fig:main across more params}) is non-trivial and specific to class-level shifts, but does not extend to corruption-based shifts where the label space is unchanged. Exploring robustness to corruption thus remains an interesting direction for future work.

\subsection{Diagnosing failures in generalization via detecting shortcut features}\label{sec:shortcut feature}
Failures in generalization are often attributed to a model specializing in its training regime. A classic example is \textit{overfitting}, where high training accuracy but low validation accuracy indicates that the model has memorized the training set rather than learned transferable patterns. Under distribution shift, however, such straightforward indicators as validation accuracy are absent.
Our findings in~\autoref{sec:prognostic results} suggest using $D_\eff$ and $\Psi_\eff$ measured on ID object manifolds as prognostics for indicating potential failure in OOD datasets with new classes of images.

Failures in OOD generalization are often attributed to reliance on \textit{shortcut} or \textit{spurious} features~\citep{geirhos2020shortcut,sagawa2020investigation,singla2021salient,yang2022understanding}. A network may correctly classify cows in typical training images, yet fail when cows appear in unusual contexts such as beaches or mountains, suggesting that background cues like grass had been used as unintended predictors of class identity~\citep{beery2018recognition}. Features such as ``grass'' correspond to microscopic details, whereas generalization performance is a macroscopic outcome. 
Effective geometric measures act as \textit{mesoscopic descriptors}, bridging how microscopic features are at play and how efficiently they are used for macroscopic behavior, such as separability~\citep{chou2025featurelearninglazyrichdichotomy} (see also~\autoref{tab:effective geometry}). 
Low $D_\eff$ and $\Psi_\eff$ indicate that the model relies on a smaller set of features, used inefficiently for separability, agreeing with shortcut-learning interpretations.

Finally, we remark that although untrained or randomly initialized networks also exhibit very high manifold dimension and poor generalization~\citep{chou2025featurelearninglazyrichdichotomy} (i.e., lazy learning, Figure 7), our analysis concerns models with comparable ID validation accuracy—i.e., after meaningful feature learning has taken place. In this regime, larger $D_\eff$ and $\Psi_\eff$ reflect richer task-relevant variability, whereas excessive compression signals overspecialization to the ID distribution.

\section{Applications to Predicting Performance of Transfer Learning}\label{sec:application}
A common scenario in applied machine learning involves selecting a pretrained model from a public repository like PyTorch Hub or Hugging Face. For a given architecture, multiple sets of weights are often available, each trained with different optimization recipes, regularization schemes, or data preprocessing pipelines. The standard heuristic is to choose the model with the highest reported in-distribution (ID) accuracy. However, it is unclear whether this metric reliably predicts performance on other downstream tasks, especially under the distribution shifts inherent in transfer learning.

Here, we apply the prognostic indicators discovered in our exploratory experiments (\autoref{sec:prognostic}) to this practical challenge. Our findings suggest a clear guiding principle for model selection: when faced with multiple weights for the same architecture, \textbf{prefer the model that exhibits higher effective manifold dimensionality ($D_\eff$) and utility ($\Psi_\eff$) on its ID data}, as this signals a greater potential for robust OOD generalization.

\begin{figure}[h]
    \centering
    \includegraphics[width=0.9\linewidth]{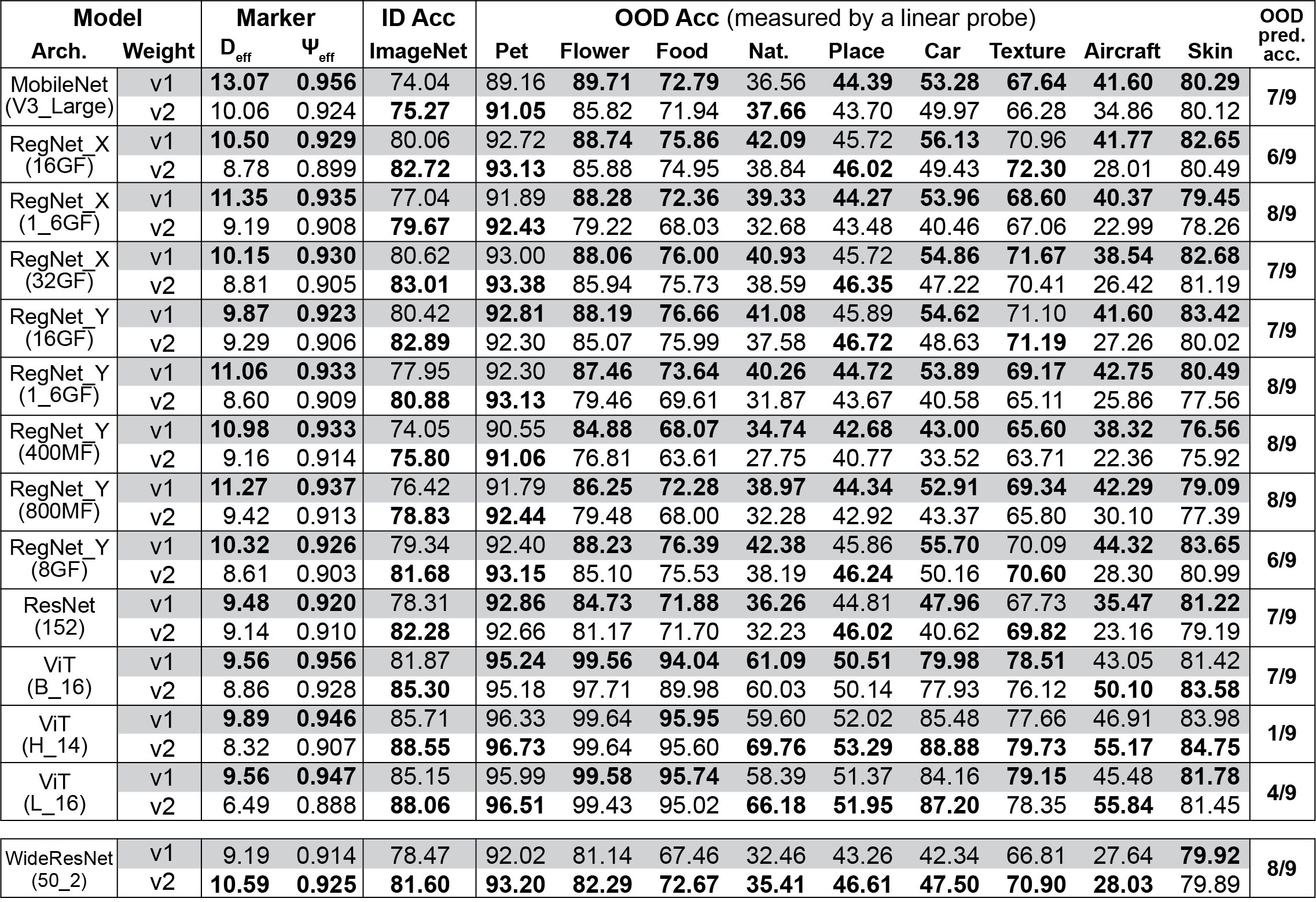}
    \caption{Predict OOD transfer performance on ImageNet-pretrained models via $D_\eff$ and $\Psi_\eff$. For the first block of models, our prognostic indicators predicted that v1 would outperform v2. For the second block of models, our prognostic indicators predicted the other way around.}
    \label{tab:pretrained}
    \vspace{-1mm}
\end{figure}

\vspace{-3mm}
\paragraph{Experimental procedure.}
To test this principle, we analyzed 20 popular architectures from PyTorch's official repository, each released with two sets of weights (v1 and v2). By design, the v2 weights achieve higher accuracy on the ID ImageNet benchmark. However, the specific changes in training procedure are often opaque to the end-user (see Table~\ref{tab:pretrained} for key differences). This heterogeneity makes for a challenging and realistic testbed for our diagnostic framework. 
For each v1/v2 pair, we first measured the $D_\eff$ and $\Psi_\eff$ of their ImageNet object manifolds. We then evaluated their OOD transfer performance on 9 image classification datasets: Flowers102~\citep{nilsback2008automated}, Stanford Cars~\citep{krause2013stanfordcars}, Places365~\citep{zhou2017places}, Food101~\citep{bossard14}, Oxford-IIIT Pet~\citep{parkhi12a}, etc. For each OOD dataset, we train a linear probe on the training set of the OOD dataset, and report the test accuracy (see~\autoref{sec:method} for details).
See~\autoref{app:applications} for more experimental details.

\vspace{-3mm}
\paragraph{Diagnosing transferability through ID effective manifold geometry.}
Consistent with the hypothesis derived from our initial explorations, we found that models with higher $D_\eff$ and $\Psi_\eff$ often demonstrated stronger OOD transfer performance, even when their ID ImageNet accuracy was lower. As shown in \autoref{tab:pretrained}, across the 20 architectures we examined, our prognostic indicators predicted that v1 would outperform v2 on OOD transfer in 14 cases (despite v2 having higher ID accuracy), that v2 would outperform v1 in 1 case, and yielded no clear verdict for the remainder. Among these 15 models and 9 OOD datasets, our prediction accuracy is 73.02\% (92 out of 126). This is much higher than using ID test accuracy as a predictor for OOD performance (37.22\%).  We remark that using some of the other markers (e.g., Neural Collapse, Participation Ratio) also yields non-trivial prediction accuracy in OOD performance. See~\autoref{sec:pretrained prognstic pred} for more results.

\vspace{-3mm}
\paragraph{Revealing differences in fine-tuning dynamics.}
Finally, we explored whether these initial feature advantages persist during full-model fine-tuning. As expected from prior work showing that the benefits of pretraining diminish with longer fine-tuning~\citep{kornblith2019betterimagenetmodelstransfer, he2018rethinkingimagenetpretraining}, both v1 and v2 initializations ultimately converged to a similar performance level. However, we observed a drastic difference in the early fine-tuning stages: models initialized with v1 weights sometimes exhibited faster learning, hinting that their features may provide a more efficient transferable starting point (Figures~\ref{fig:finetuning_flowers102},~\ref{fig:finetuning_stanfordcars}). These results show that test-relevant geometric measures can reveal differences in fine-tuning dynamics, motivating future study on their role in transfer learning.

\section{Discussion}
We introduced a diagnostic, system-level paradigm for anticipating generalization failure in neural networks. Instead of reconstructing detailed internal mechanisms, we treated task-relevant geometric markers of ID representations as prognostic indicators. Through discovering prognostic markers in medium-sized experiments, we found that overcompression of object manifold dimension consistently predicts failures in OOD generalization. Applied to ImageNet-pretrained models—a far more heterogeneous real-world setting—our prognostic measures predict which models transfer more robustly across tasks. Together, these results demonstrate the power of a diagnostic framework for studying generalization failures. This work opens up several future directions. 
\begin{itemize}[leftmargin=*]
\item \textbf{Theoretical foundations.} 
In~\autoref{sec:shortcut feature}, we link overcompression of object manifolds to overspecialization of learned features. Strengthening the theoretical basis of this hypothesis is important. A related question is whether incorrectly classified OOD examples share common traits that can be explained by the overspecialization intuition.
\item \textbf{Causal mechanisms and interventions.} Geometric indicators could inspire investigation into underlying causal mechanisms and practical interventions, such as geometry-aware regularization, early-stopping criteria, or model selection rules that prioritize robustness alongside accuracy.
\item \textbf{Extending the proposed diagnostic research framework.} Expanding our proposed analysis framework beyond vision to language, reinforcement learning, or multi-modal models remains an open challenge. A natural starting point is to first identify and characterize the relevant failure modes in each domain, and then examine how representational markers correlate with those failures. Another direction is to extend our findings into deployable protocols for diagnosing OOD failures across a wider range of models and datasets.
\item \textbf{Linking diagnostics to parameter transfer.} A future direction is to explore whether insights from our controlled experiments can inform parameter transfer between models of different scales, as in Net2Net~\citep{chen2015net2net}. While our focus here is on diagnostics, connecting to weight transfer could provide a complementary path for robust initialization.
\item \textbf{Parallels with neuroscience.} High-dimensional yet structured codes in the brain have been linked to generalization in neuroscience studies. Our hypothesis linking manifold compression to feature overspecialization may provide a framework for interpreting these findings and exploring common principles across biological and artificial systems.
\end{itemize}

Theoretical work on neural networks has long been shaped by mathematics and physics, with an emphasis on bottom-up mechanistic explanations. We suggest that the history of medicine offers a complementary perspective: effective diagnostics can anticipate risks and guide treatment well before underlying causal mechanisms are fully understood. Neural networks, as emergent high-dimensional systems, may likewise benefit from a diagnostic science that anticipates vulnerabilities and guides future mechanistic insight.

\subsubsection*{Acknowledgments}
We thank Hang Le for the helpful discussion. This work was supported by the Center for Computational Neuroscience at the Flatiron Institute, Simons Foundation. S.C. was partially supported by a Sloan Research Fellowship, a Klingenstein-Simons Award, and the Samsung Advanced Institute of Technology project, “Next Generation Deep Learning: From Pattern Recognition to AI.” All experiments were performed using the Flatiron Institute’s high-performance computing cluster.
Yao-Yuan Yang worked in an advisory capacity.

\subsubsection*{Code Availability}
All code required to reproduce the figures presented is available under an MIT License at \href{https://github.com/chung-neuroai-lab/ood-generalization-geometry}{https://github.com/chung-neuroai-lab/ood-generalization-geometry}

\bibliography{iclr2026_conference}

@misc{sagawa2020distributionallyrobustneuralnetworks,
      title={Distributionally Robust Neural Networks for Group Shifts: On the Importance of Regularization for Worst-Case Generalization}, 
      author={Shiori Sagawa and Pang Wei Koh and Tatsunori B. Hashimoto and Percy Liang},
      year={2020},
      eprint={1911.08731},
      archivePrefix={arXiv},
      primaryClass={cs.LG},
      url={https://arxiv.org/abs/1911.08731}, 
}

@article{liu2021towards,
  title={Towards out-of-distribution generalization: A survey},
  author={Liu, Jiashuo and Shen, Zheyan and He, Yue and Zhang, Xingxuan and Xu, Renzhe and Yu, Han and Cui, Peng},
  journal={arXiv preprint arXiv:2108.13624},
  year={2021}
}

@article{yang2024generalized,
  title={Generalized out-of-distribution detection: A survey},
  author={Yang, Jingkang and Zhou, Kaiyang and Li, Yixuan and Liu, Ziwei},
  journal={International Journal of Computer Vision},
  volume={132},
  number={12},
  pages={5635--5662},
  year={2024},
  publisher={Springer}
}

@article{chou2025glue,
  title={Geometry Linked to Untangling Efficiency Reveals Structure and Computation in Neural Populations},
  author={Chou, Chi-Ning and Kim, Royoung and Arend, Luke and Yang, Yao-Yuan and Mensh, Brett and Shim, Won Mok and Perich, Matthew and Chung, SueYeon},
  journal={bioRxiv},
  elocation-id = {2024.02.26.582157},
  doi = {10.1101/2024.02.26.582157},
  year={2025},
  publisher={Cold Spring Harbor Laboratory}
}

@article{chung2018classification,
  title={Classification and geometry of general perceptual manifolds},
  author={Chung, SueYeon and Lee, Daniel D and Sompolinsky, Haim},
  journal={Physical Review X},
  year={2018},
  publisher={APS}
}

@article{wakhloo2023linear,
  title={Linear classification of neural manifolds with correlated variability},
  author={Wakhloo, Albert J and Sussman, Tamara J and Chung, SueYeon},
  journal={Physical Review Letters},
  year={2023},
  publisher={APS}
}

@article{cohen2020separability,
  title={Separability and geometry of object manifolds in deep neural networks},
  author={Cohen, Uri and Chung, SueYeon and Lee, Daniel D and Sompolinsky, Haim},
  journal={Nature communications},
  year={2020},
  publisher={Nature Publishing Group UK London}
}

@article{paraouty2023sensory,
  title={Sensory cortex plasticity supports auditory social learning},
  author={Paraouty, Nihaad and Yao, Justin D and Varnet, L{\'e}o and Chou, Chi-Ning and Chung, SueYeon and Sanes, Dan H},
  journal={Nature communications},
  volume={14},
  number={1},
  pages={5828},
  year={2023},
  publisher={Nature Publishing Group UK London}
}

@inproceedings{kuoch2024probing,
  title={Probing Biological and Artificial Neural Networks with Task-dependent Neural Manifolds},
  author={Kuoch, Michael and Chou, Chi-Ning and Parthasarathy, Nikhil and Dapello, Joel and DiCarlo, James J and Sompolinsky, Haim and Chung, SueYeon},
  booktitle={Conference on Parsimony and Learning (Proceedings Track)},
  year={2024}
}

@article{yao2023transformation,
  title={Transformation of acoustic information to sensory decision variables in the parietal cortex},
  author={Yao, Justin D and Zemlianova, Klavdia O and Hocker, David L and Savin, Cristina and Constantinople, Christine M and Chung, SueYeon and Sanes, Dan H},
  journal={Proceedings of the National Academy of Sciences},
  volume={120},
  number={2},
  pages={e2212120120},
  year={2023},
  publisher={National Acad Sciences}
}

@inproceedings{mamou2020emergence,
  title={Emergence of separable manifolds in deep language representations},
  author={Mamou, Jonathan and Le, Hang and Del Rio, Miguel A and Stephenson, Cory and Tang, Hanlin and Kim, Yoon and Chung, SueYeon},
  booktitle={Proceedings of the 37th International Conference on Machine Learning},
  pages={6713--6723},
  year={2020}
}

@article{stephenson2021geometry,
  title={On the geometry of generalization and memorization in deep neural networks},
  author={Stephenson, Cory and Padhy, Suchismita and Ganesh, Abhinav and Hui, Yue and Tang, Hanlin and Chung, SueYeon},
  journal={arXiv preprint arXiv:2105.14602},
  year={2021}
}

@inproceedings{he2016deep,
  title={Deep residual learning for image recognition},
  author={He, Kaiming and Zhang, Xiangyu and Ren, Shaoqing and Sun, Jian},
  booktitle={Proceedings of the IEEE conference on computer vision and pattern recognition},
  pages={770--778},
  year={2016}
}

@article{yang2022understanding,
  title={Understanding rare spurious correlations in neural networks},
  author={Yang, Yao-Yuan and Chou, Chi-Ning and Chaudhuri, Kamalika},
  journal={arXiv preprint arXiv:2202.05189},
  year={2022}
}

@article{singla2021salient,
  title={Salient imagenet: How to discover spurious features in deep learning?},
  author={Singla, Sahil and Feizi, Soheil},
  journal={arXiv preprint arXiv:2110.04301},
  year={2021}
}

@inproceedings{sagawa2020investigation,
  title={An investigation of why overparameterization exacerbates spurious correlations},
  author={Sagawa, Shiori and Raghunathan, Aditi and Koh, Pang Wei and Liang, Percy},
  booktitle={International Conference on Machine Learning},
  pages={8346--8356},
  year={2020},
  organization={PMLR}
}

@article{gardner1988optimal,
  title={Optimal storage properties of neural network models},
  author={Gardner, Elizabeth and Derrida, Bernard},
  journal={Journal of Physics A: Mathematical and general},
  volume={21},
  number={1},
  pages={271},
  year={1988},
  publisher={IOP Publishing}
}

@article{ansuini2019intrinsic,
  title={Intrinsic dimension of data representations in deep neural networks},
  author={Ansuini, Alessio and Laio, Alessandro and Macke, Jakob H and Zoccolan, Davide},
  journal={Advances in Neural Information Processing Systems},
  volume={32},
  year={2019}
}

@inproceedings{hendrycks2018benchmarking,
  title={Benchmarking Neural Network Robustness to Common Corruptions and Perturbations},
  author={Hendrycks, Dan and Dietterich, Thomas},
  booktitle={International Conference on Learning Representations},
  year={2018}
}

@inproceedings{simonyan2015very,
  title={Very deep convolutional networks for large-scale image recognition},
  author={Simonyan, K and Zisserman, A},
  booktitle={3rd International Conference on Learning Representations (ICLR 2015)},
  year={2015},
  organization={Computational and Biological Learning Society}
}

@article{vgg,
  title={Very deep convolutional networks for large-scale image recognition},
  author={Simonyan, Karen and Zisserman, Andrew},
  journal={arXiv preprint arXiv:1409.1556},
  year={2014}
}

@article{alain2016understanding,
  title={Understanding intermediate layers using linear classifier probes},
  author={Alain, Guillaume and Bengio, Yoshua},
  journal={arXiv preprint arXiv:1610.01644},
  year={2016}
}

@article{olah2017feature,
  author = {Olah, Chris and Mordvintsev, Alexander and Schubert, Ludwig},
  title = {Feature Visualization},
  journal = {Distill},
  year = {2017},
  note = {https://distill.pub/2017/feature-visualization},
  doi = {10.23915/distill.00007}
}

@article{chung2021neural,
  title={Neural population geometry: An approach for understanding biological and artificial neural networks},
  author={Chung, SueYeon and Abbott, Larry F},
  journal={Current opinion in neurobiology},
  volume={70},
  pages={137--144},
  year={2021},
  publisher={Elsevier}
}

@book{vershynin2018high,
  title={High-dimensional probability: An introduction with applications in data science},
  author={Vershynin, Roman},
  volume={47},
  year={2018},
  publisher={Cambridge university press}
}

@inproceedings{kirsanov2025,
    title = "The Geometry of Prompting: Unveiling Distinct Mechanisms of Task Adaptation in Language Models",
    author = "Kirsanov, Artem  and
      Chou, Chi-Ning  and
      Cho, Kyunghyun  and
      Chung, SueYeon",
    booktitle = "Findings of the Association for Computational Linguistics: NAACL 2025",
    month = apr,
    year = "2025",
    publisher = "Association for Computational Linguistics",
    pages = "1855--1888",
}

@article{hu2024representational,
  title={Representational learning by optimization of neural manifolds in an olfactory memory network},
  author={Hu, Bo and Temiz, Nesibe Z and Chou, Chi-Ning and Rupprecht, Peter and Meissner-Bernard, Claire and Titze, Benjamin and Chung, SueYeon and Friedrich, Rainer W},
  journal={bioRxiv},
  pages={2024--11},
  year={2024},
  publisher={Cold Spring Harbor Laboratory}
}

@inproceedings{chou2025featurelearninglazyrichdichotomy,
title={Feature Learning beyond the Lazy-Rich Dichotomy: Insights from Representational Geometry},
author={Chou, Chi-Ning and Le, Hang and Wang, Yichen and Chung, SueYeon},
booktitle={Forty-second International Conference on Machine Learning},
year={2025},
}

@inproceedings{
loshchilov2018decoupled,
title={Decoupled Weight Decay Regularization},
author={Ilya Loshchilov and Frank Hutter},
booktitle={International Conference on Learning Representations},
year={2019},
}

@misc{krizhevsky2009learning,
  title={Learning multiple layers of features from tiny images},
  author={Krizhevsky, Alex and Hinton, Geoffrey and others},
  year={2009},
  publisher={Toronto, ON, Canada}
}

@InProceedings{pmlr-v97-tan19a,
  title = 	 {{E}fficient{N}et: Rethinking Model Scaling for Convolutional Neural Networks},
  author =       {Tan, Mingxing and Le, Quoc},
  booktitle = 	 {Proceedings of the 36th International Conference on Machine Learning},
  year = 	 {2019},
  editor = 	 {Chaudhuri, Kamalika and Salakhutdinov, Ruslan},
  series = 	 {Proceedings of Machine Learning Research},
  publisher =    {PMLR},
}

@article{howard2017mobilenets,
  title={Mobilenets: Efficient convolutional neural networks for mobile vision applications},
  author={Howard, Andrew G and Zhu, Menglong and Chen, Bo and Kalenichenko, Dmitry and Wang, Weijun and Weyand, Tobias and Andreetto, Marco and Adam, Hartwig},
  journal={arXiv preprint arXiv:1704.04861},
  year={2017}
}

@inproceedings{huang2017densely,
  title={Densely connected convolutional networks},
  author={Huang, Gao and Liu, Zhuang and Van Der Maaten, Laurens and Weinberger, Kilian Q},
  booktitle={Proceedings of the IEEE conference on computer vision and pattern recognition},
  pages={4700--4708},
  year={2017}
}

@article{zhu2023variance,
  title={Variance-covariance regularization improves representation learning},
  author={Zhu, Jiachen and Evtimova, Katrina and Chen, Yubei and Shwartz-Ziv, Ravid and LeCun, Yann},
  journal={arXiv preprint arXiv:2306.13292},
  year={2023}
}

@inproceedings{morcos2018importance,
  title={On the importance of single directions for generalization},
  author={Morcos, Ari S and Barrett, David GT and Rabinowitz, Neil C and Botvinick, Matthew},
  booktitle={International Conference on Learning Representations},
  year={2018}
}

@article{stringer2019high,
  title={High-dimensional geometry of population responses in visual cortex},
  author={Stringer, Carsen and Pachitariu, Marius and Steinmetz, Nicholas and Carandini, Matteo and Harris, Kenneth D},
  journal={Nature},
  volume={571},
  number={7765},
  pages={361--365},
  year={2019},
  publisher={Nature Publishing Group UK London}
}

@inproceedings{deng2009imagenet,
  title={Imagenet: A large-scale hierarchical image database},
  author={Deng, Jia and Dong, Wei and Socher, Richard and Li, Li-Jia and Li, Kai and Fei-Fei, Li},
  booktitle={2009 IEEE conference on computer vision and pattern recognition},
  pages={248--255},
  year={2009},
  organization={Ieee}
}

@inproceedings{krause2013stanfordcars,
title = {3D Object Representations for Fine-Grained Categorization},
booktitle={Proceedings of the IEEE international conference on computer vision workshops},
author = {Krause, Jonathan and Stark, Michael and Deng, Jia and Fei-Fei, Li},
year = {2013},
}

@article{barak2013sparseness,
  title={The sparseness of mixed selectivity neurons controls the generalization--discrimination trade-off},
  author={Barak, Omri and Rigotti, Mattia and Fusi, Stefano},
  journal={Journal of Neuroscience},
  volume={33},
  number={9},
  pages={3844--3856},
  year={2013},
  publisher={Society for Neuroscience}
}

@article{mastrogiuseppe2018linking,
  title={Linking connectivity, dynamics, and computations in low-rank recurrent neural networks},
  author={Mastrogiuseppe, Francesca and Ostojic, Srdjan},
  journal={Neuron},
  volume={99},
  number={3},
  pages={609--623},
  year={2018},
  publisher={Elsevier}
}

@article{rigotti2013importance,
  title={The importance of mixed selectivity in complex cognitive tasks},
  author={Rigotti, Mattia and Barak, Omri and Warden, Melissa R and Wang, Xiao-Jing and Daw, Nathaniel D and Miller, Earl K and Fusi, Stefano},
  journal={Nature},
  volume={497},
  number={7451},
  pages={585--590},
  year={2013},
  publisher={Nature Publishing Group UK London}
}

@misc{yun2023transformervisualizationdictionarylearning,
      title={Transformer visualization via dictionary learning: contextualized embedding as a linear superposition of transformer factors}, 
      author={Zeyu Yun and Yubei Chen and Bruno A Olshausen and Yann LeCun},
      year={2023},
      eprint={2103.15949},
      archivePrefix={arXiv},
      primaryClass={cs.CL},
      url={https://arxiv.org/abs/2103.15949}, 
}

@misc{cunningham2023sparseautoencodershighlyinterpretable,
      title={Sparse Autoencoders Find Highly Interpretable Features in Language Models}, 
      author={Hoagy Cunningham and Aidan Ewart and Logan Riggs and Robert Huben and Lee Sharkey},
      year={2023},
      eprint={2309.08600},
      archivePrefix={arXiv},
      primaryClass={cs.LG},
      url={https://arxiv.org/abs/2309.08600}, 
}

@article{dunefsky2024transcoders,
  title={Transcoders find interpretable llm feature circuits},
  author={Dunefsky, Jacob and Chlenski, Philippe and Nanda, Neel},
  journal={Advances in Neural Information Processing Systems},
  volume={37},
  pages={24375--24410},
  year={2024}
}

@inproceedings{Olah2020ZoomIA,
  title={Zoom In: An Introduction to Circuits},
  author={Christopher Olah and Nick Cammarata and Ludwig Schubert and Gabriel Goh and Michael Petrov and Shan Carter},
  year={2020},
  url={https://api.semanticscholar.org/CorpusID:215930358}
}

@article{zhou2017places,
  title={Places: A 10 million Image Database for Scene Recognition},
  author={Zhou, Bolei and Lapedriza, Agata and Khosla, Aditya and Oliva, Aude and Torralba, Antonio},
  journal={IEEE Transactions on Pattern Analysis and Machine Intelligence},
  year={2017},
  publisher={IEEE}
}

@inproceedings{nilsback2008automated,
  title={Automated flower classification over a large number of classes},
  author={Nilsback, Maria-Elena and Zisserman, Andrew},
  booktitle={2008 Sixth Indian conference on computer vision, graphics \& image processing},
  pages={722--729},
  year={2008},
  organization={IEEE}
}

@inproceedings{radosavovic2020designing,
  title={Designing network design spaces},
  author={Radosavovic, Ilija and Kosaraju, Raj Prateek and Girshick, Ross and He, Kaiming and Doll{\'a}r, Piotr},
  booktitle={Proceedings of the IEEE/CVF conference on computer vision and pattern recognition},
  pages={10428--10436},
  year={2020}
}

@article{zagoruyko2016wide,
  title={Wide residual networks},
  author={Zagoruyko, Sergey and Komodakis, Nikos},
  journal={arXiv preprint arXiv:1605.07146},
  year={2016}
}

@article{geiger2025causal,
  title={Causal abstraction: A theoretical foundation for mechanistic interpretability},
  author={Geiger, Atticus and Ibeling, Duligur and Zur, Amir and Chaudhary, Maheep and Chauhan, Sonakshi and Huang, Jing and Arora, Aryaman and Wu, Zhengxuan and Goodman, Noah and Potts, Christopher and others},
  journal={Journal of Machine Learning Research},
  volume={26},
  number={83},
  pages={1--64},
  year={2025}
}

@article{mueller2024quest,
  title={The quest for the right mediator: A history, survey, and theoretical grounding of causal interpretability},
  author={Mueller, Aaron and Brinkmann, Jannik and Li, Millicent and Marks, Samuel and Pal, Koyena and Prakash, Nikhil and Rager, Can and Sankaranarayanan, Aruna and Sharma, Arnab Sen and Sun, Jiuding and others},
  journal={arXiv preprint arXiv:2408.01416},
  year={2024}
}

@article{li2024representations,
  title={Representations and generalization in artificial and brain neural networks},
  author={Li, Qianyi and Sorscher, Ben and Sompolinsky, Haim},
  journal={Proceedings of the National Academy of Sciences},
  volume={121},
  number={27},
  pages={e2311805121},
  year={2024},
  publisher={National Academy of Sciences}
}

@article{papyan2020prevalence,
  title={Prevalence of neural collapse during the terminal phase of deep learning training},
  author={Papyan, Vardan and Han, XY and Donoho, David L},
  journal={Proceedings of the National Academy of Sciences},
  volume={117},
  number={40},
  pages={24652--24663},
  year={2020},
  publisher={National Academy of Sciences}
}

@inproceedings{galanti2022on,
title={On the Role of Neural Collapse in Transfer Learning},
author={Tomer Galanti and Andr{\'a}s Gy{\"o}rgy and Marcus Hutter},
booktitle={International Conference on Learning Representations},
year={2022},
url={https://openreview.net/forum?id=SwIp410B6aQ}
}

@article{d2022underspecification,
  title={Underspecification presents challenges for credibility in modern machine learning},
  author={D'Amour, Alexander and Heller, Katherine and Moldovan, Dan and Adlam, Ben and Alipanahi, Babak and Beutel, Alex and Chen, Christina and Deaton, Jonathan and Eisenstein, Jacob and Hoffman, Matthew D and others},
  journal={Journal of Machine Learning Research},
  volume={23},
  number={226},
  pages={1--61},
  year={2022}
}

@article{cortes1995support,
  title={Support-vector networks},
  author={Cortes, Corinna and Vapnik, Vladimir},
  journal={Machine learning},
  volume={20},
  number={3},
  pages={273--297},
  year={1995},
  publisher={Springer}
}

@article{geirhos2020shortcut,
  title={Shortcut learning in deep neural networks},
  author={Geirhos, Robert and Jacobsen, J{\"o}rn-Henrik and Michaelis, Claudio and Zemel, Richard and Brendel, Wieland and Bethge, Matthias and Wichmann, Felix A},
  journal={Nature Machine Intelligence},
  volume={2},
  number={11},
  pages={665--673},
  year={2020},
  publisher={Nature Publishing Group UK London}
}

@inproceedings{beery2018recognition,
  title={Recognition in terra incognita},
  author={Beery, Sara and Van Horn, Grant and Perona, Pietro},
  booktitle={Proceedings of the European conference on computer vision (ECCV)},
  pages={456--473},
  year={2018}
}

@inproceedings{park2024linear,
  title={The Linear Representation Hypothesis and the Geometry of Large Language Models},
  author={Park, Kiho and Choe, Yo Joong and Veitch, Victor},
  booktitle={International Conference on Machine Learning},
  pages={39643--39666},
  year={2024},
  organization={PMLR}
}

@article{mignacco2025nonlinear,
  title={Nonlinear classification of neural manifolds with contextual information},
  author={Mignacco, Francesca and Chou, Chi-Ning and Chung, SueYeon},
  journal={Physical Review E},
  volume={111},
  number={3},
  pages={035302},
  year={2025},
  publisher={APS}
}

@misc{kornblith2019betterimagenetmodelstransfer,
      title={Do Better ImageNet Models Transfer Better?}, 
      author={Simon Kornblith and Jonathon Shlens and Quoc V. Le},
      year={2019},
      eprint={1805.08974},
      archivePrefix={arXiv},
      primaryClass={cs.CV},
      url={https://arxiv.org/abs/1805.08974}, 
}

@misc{he2018rethinkingimagenetpretraining,
      title={Rethinking ImageNet Pre-training}, 
      author={Kaiming He and Ross Girshick and Piotr Dollár},
      year={2018},
      eprint={1811.08883},
      archivePrefix={arXiv},
      primaryClass={cs.CV},
      url={https://arxiv.org/abs/1811.08883}, 
}

@misc{xie2017aggregatedresidualtransformationsdeep,
      title={Aggregated Residual Transformations for Deep Neural Networks}, 
      author={Saining Xie and Ross Girshick and Piotr Dollár and Zhuowen Tu and Kaiming He},
      year={2017},
      eprint={1611.05431},
      archivePrefix={arXiv},
      primaryClass={cs.CV},
      url={https://arxiv.org/abs/1611.05431}, 
}

@article{chen2015net2net,
  title={Net2net: Accelerating learning via knowledge transfer},
  author={Chen, Tianqi and Goodfellow, Ian and Shlens, Jonathon},
  journal={arXiv preprint arXiv:1511.05641},
  year={2015}
}

@inproceedings{bossard14,
  title = {Food-101 -- Mining Discriminative Components with Random Forests},
  author = {Bossard, Lukas and Guillaumin, Matthieu and Van Gool, Luc},
  booktitle = {European Conference on Computer Vision},
  year = {2014}
}

@InProceedings{parkhi12a,
  author       = "Omkar M. Parkhi and Andrea Vedaldi and Andrew Zisserman and C. V. Jawahar",
  title        = "Cats and Dogs",
  booktitle    = "IEEE Conference on Computer Vision and Pattern Recognition",
  year         = "2012",
}

@inproceedings{guillory2021predicting,
  title={Predicting with confidence on unseen distributions},
  author={Guillory, Devin and Shankar, Vaishaal and Ebrahimi, Sayna and Darrell, Trevor and Schmidt, Ludwig},
  booktitle={Proceedings of the IEEE/CVF international conference on computer vision},
  pages={1134--1144},
  year={2021}
}

@article{ammar2023neco,
  title={Neco: Neural collapse based out-of-distribution detection},
  author={Ammar, Mou{\"\i}n Ben and Belkhir, Nacim and Popescu, Sebastian and Manzanera, Antoine and Franchi, Gianni},
  journal={arXiv preprint arXiv:2310.06823},
  year={2023}
}

@article{liu2020energy,
  title={Energy-based out-of-distribution detection},
  author={Liu, Weitang and Wang, Xiaoyun and Owens, John and Li, Yixuan},
  journal={Advances in neural information processing systems},
  volume={33},
  pages={21464--21475},
  year={2020}
}

@article{masarczyk2023tunnel,
  title={The tunnel effect: Building data representations in deep neural networks},
  author={Masarczyk, Wojciech and Ostaszewski, Mateusz and Imani, Ehsan and Pascanu, Razvan and Mi{\l}o{\'s}, Piotr and Trzcinski, Tomasz},
  journal={Advances in Neural Information Processing Systems},
  volume={36},
  pages={76772--76805},
  year={2023}
}

@article{harun2024variables,
  title={What variables affect out-of-distribution generalization in pretrained models?},
  author={Harun, Yousuf and Lee, Kyungbok and Gallardo, Jhair and Krishnan, Giri and Kanan, Christopher},
  journal={Advances in Neural Information Processing Systems},
  volume={37},
  pages={56479--56525},
  year={2024}
}

@inproceedings{haruncontrolling,
  title={Controlling Neural Collapse Enhances Out-of-Distribution Detection and Transfer Learning},
  author={Harun, Md Yousuf and Gallardo, Jhair and Kanan, Christopher},
  booktitle={Forty-second International Conference on Machine Learning},
  year={2025},
}

@techreport{maji13fine-grained,
  title = {Fine-Grained Visual Classification of Aircraft},
  author = {S. Maji and J. Kannala and E. Rahtu and M. Blaschko and A. Vedaldi},
  year = {2013},
  archivePrefix = {arXiv},
  eprint = {1306.5151},
  primaryClass = "cs-cv",
}

@inproceedings{cimpoi14describing,
  author = {M. Cimpoi and S. Maji and I. Kokkinos and S. Mohamed and and A. Vedaldi},
  title = {Describing Textures in the Wild},
  booktitle = {Proceedings of the {IEEE} Conf. on Computer Vision and Pattern Recognition ({CVPR})},
  year = {2014}
}

@inproceedings{van2018inaturalist,
  title={The inaturalist species classification and detection dataset},
  author={Van Horn, Grant and Mac Aodha, Oisin and Song, Yang and Cui, Yin and Sun, Chen and Shepard, Alex and Adam, Hartwig and Perona, Pietro and Belongie, Serge},
  booktitle={Proceedings of the IEEE conference on computer vision and pattern recognition},
  pages={8769--8778},
  year={2018}
}

@article{tschandl2018ham10000,
  title={The HAM10000 dataset, a large collection of multi-source dermatoscopic images of common pigmented skin lesions},
  author={Tschandl, Philipp and Rosendahl, Cliff and Kittler, Harald},
  journal={Scientific data},
  volume={5},
  number={1},
  pages={1--9},
  year={2018},
  publisher={Nature Publishing Group}
}

@inproceedings{meloux2025everything,
  title={Everything, everywhere, all at once: is mechanistic interpretability identifiable?},
  author={M{\'e}loux, Maxime and Maniu, Silviu and Portet, Fran{\c{c}}ois and Peyrard, Maxime},
  booktitle={The Thirteenth International Conference on Learning Representations},
  year={2025}
}

@article{li2025can,
  title={Can Interpretation Predict Behavior on Unseen Data?},
  author={Li, Victoria R and Kaufmann, Jenny and Wattenberg, Martin and Alvarez-Melis, David and Saphra, Naomi},
  journal={arXiv preprint arXiv:2507.06445},
  year={2025}
}
\bibliographystyle{iclr2026_conference}

\newpage
\appendix
\section{Experimental Settings}\label{app:experiment}
In this section, we provide a complete description of our experimental setup to facilitate reproducibility.


\subsection{Datasets}
Our study utilized a range of standard image classification datasets, which served different roles: either as in-distribution (ID) training sources or out-of-distribution (OOD) evaluation benchmarks across two distinct experimental settings. \autoref{tab:dataset_summary} provides a summary of these roles. Below, we describe each dataset and the specific preprocessing pipelines applied.

\begin{table}[h!]
\caption{Summary of dataset roles in our experiments.}
\label{tab:dataset_summary}
\centering
\footnotesize
\renewcommand{\arraystretch}{1.2}
\begin{tabularx}{\linewidth}{ l >{\raggedright\arraybackslash}X >{\raggedright\arraybackslash}X }
\toprule 
\textbf{Experimental Setting} & \textbf{In-Distribution (ID) Dataset} & \textbf{Out-of-Distribution (OOD) Datasets} \\ 
\midrule 
\textbf{Prognostic discovery} (\autoref{sec:prognostic}) & 
CIFAR-10 & 
CIFAR-100 \newline ImageNet-1k (resized to 32x32) \\ 
\addlinespace 
\textbf{Transfer Learning Applications} (\autoref{sec:application}) & 
ImageNet-1k & 
Flowers102 \newline Stanford Cars \newline Places365 \newline Oxford-IIIT Pets \newline Food-101 \newline iNaturalist 2018 \newline DTD \newline FGVC-Aircraft \newline HAM10000 \\ 
\bottomrule 
\end{tabularx}
\end{table}

\paragraph{Datasets for prognostic discovery.}
In our controlled medium-scale experiments, we trained models from scratch on a single ID dataset and evaluated their generalization to two different OOD datasets with disjoint classes.

\begin{itemize}[leftmargin=*]
\item \textbf{CIFAR-10}~\citep{krizhevsky2009learning} served as our primary \textbf{in-distribution (ID)} dataset for training. It contains 60,000 color images of $32\times32$ pixels, split into 50,000 training and 10,000 test images across 10 object categories. For training, we normalized images using a per-channel mean of $(0.4914, 0.4822, 0.4465)$ and a standard deviation of $(0.2023, 0.1994, 0.2010)$. We also applied standard data augmentation: padding with 4 pixels on each side, followed by a random $32\times32$ crop and a random horizontal flip with 50\% probability. For evaluating ID test accuracy, augmentation was disabled.

\item \textbf{CIFAR-100}~\citep{krizhevsky2009learning} was used as the primary \textbf{out-of-distribution (OOD)} benchmark. It has the same image format and size as CIFAR-10 but contains 100 distinct object classes with no overlap. For OOD evaluation, images were only normalized using the CIFAR-10 statistics; no data augmentation was applied to ensure a deterministic evaluation protocol.

\item \textbf{ImageNet-1k}~\citep{deng2009imagenet} was used as a second, more challenging \textbf{OOD} benchmark to test generalization under a significant domain shift. This dataset contains over 1.2 million high-resolution images from 1,000 categories. To maintain compatibility with our CIFAR-trained models, all ImageNet images were resized to $32\times32$ pixels using bicubic interpolation. They were then normalized using the standard ImageNet per-channel mean $(0.485, 0.456, 0.406)$ and standard deviation $(0.229, 0.224, 0.225)$. No data augmentation was applied during evaluation.

\end{itemize}

\paragraph{Datasets for Transfer Learning Applications.}
In this setting, we analyzed publicly available models pretrained on ImageNet-1k and evaluated their transferability to three downstream, fine-grained classification tasks.

\begin{itemize}[leftmargin=*]
\item \textbf{ImageNet-1k} served as the \textbf{in-distribution (ID)} dataset, as all models we analyzed were pretrained on it. For measuring the ID geometric markers, we used the official validation set. Images were processed according to the standard pipeline for each model: resized to $256\times256$, center-cropped to $224\times224$, and normalized using the standard ImageNet mean and standard deviation.

\item \textbf{Flowers102}~\citep{nilsback2008automated} is a fine-grained OOD dataset containing 8,189 images of flowers belonging to 102 different categories.

\item \textbf{Stanford Cars}~\citep{krause2013stanfordcars} is another fine-grained OOD dataset consisting of 16,185 images of cars, categorized by 196 classes (e.g., make, model, and year).

\item \textbf{Places365}~\citep{zhou2017places} is a large-scale scene-centric OOD dataset with over 1.8 million images from 365 scene categories.

\item \textbf{Oxford-IIIT Pets}~\citep{parkhi12a} contains a 37-category pet dataset with roughly 200 images for each class.

\item \textbf{Food-101}~\citep{bossard14} includes 101,000 images of 101 food dishes (750 training and 250 test images per class). 
The dataset exhibits large variation in presentation, lighting, and style.

\item \textbf{iNaturalist 2018}~\citep{van2018inaturalist} consists of over 450,000 training images from more than 8,000 species of plants, animals, and fungi, collected and verified by citizen scientists on the iNaturalist platform. The long-tailed distribution and diverse real-world conditions make this dataset highly challenging for transfer evaluation.

\item \textbf{Describable Textures Dataset (DTD)}~\citep{cimpoi14describing} contains 5,640 texture images annotated with 47 describable texture attributes. 
Images span varied materials, lighting, and scales.

\item \textbf{FGVC-Aircraft}~\citep{maji13fine-grained} is a fine-grained visual classification dataset containing 10,000 images across 100 aircraft variants. 
Images differ in viewpoint, environment, and model-year variations.

\item \textbf{HAM10000}~\citep{tschandl2018ham10000} is a dermatology image dataset containing 10{,}015 dermatoscopic images drawn from seven diagnostic categories (e.g., melanocytic nevi, melanoma, benign keratosis, vascular lesions). The images exhibit substantial variation in acquisition conditions, anatomical location, and lesion appearance, making HAM a visually and semantically distinct OOD dataset relative to natural-image pretraining.

\end{itemize}

For all three OOD datasets in this setting, images were resized to $224 \times 224$ pixels using bicubic interpolation and then normalized. During the OOD evaluation via linear probing, no data augmentation was applied. For the full-model fine-tuning experiments (see \autoref{fig:finetuning_flowers102}), data augmentation was applied during the training phase, which included random horizontal flipping (with a 50\% probability) and color jitter. These augmentations were disabled during the evaluation of model checkpoints on the OOD validation subsets.

\subsection{Model Architectures}

\subsubsection{Models for prognostic discovery (\autoref{sec:prognostic})}

To ensure our findings generalize across different model design philosophies, our exploratory studies included a diverse set of convolutional neural network (CNN) architectures. All models were adapted for CIFAR-scale ($32 \times 32$ pixel) inputs and trained from random initialization, ensuring that their learned representations were not influenced by prior pretraining:

\begin{itemize}[leftmargin=*]
\item \textbf{ResNet}~\citep{he2016deep}: A family of foundational deep residual networks that utilize skip connections to enable effective training of very deep models. We used the ResNet-18, ResNet-34, and ResNet-50 variants.

\item \textbf{VGG}~\citep{simonyan2015very}: Classic deep feedforward networks characterized by their architectural simplicity and sequential stacking of small $3\times3$ convolutions. We included VGG-13, VGG-16, and VGG-19, each augmented with batch normalization.

\item \textbf{MobileNetV1}~\citep{howard2017mobilenets}: A lightweight architecture designed for computational efficiency through the use of depthwise separable convolutions.

\item \textbf{EfficientNet-B0}~\citep{pmlr-v97-tan19a}: A modern, highly efficient model that systematically scales network depth, width, and resolution using a compound scaling method.

\item \textbf{DenseNet}~\citep{huang2017densely}: An architecture designed to maximize feature reuse and improve gradient flow by connecting each layer to every other subsequent layer within dense blocks.
\end{itemize}

This selection spans a wide architectural landscape, including canonical residual and feedforward designs, modern efficient networks, and architectures with alternative connectivity patterns. This diversity allows us to validate that our findings are a general property of deep representations, rather than an artifact of a specific model family.

\subsubsection{Models for Transfer Learning Applications (\autoref{sec:application})}

For the experiments in Section \ref{sec:application}, we shifted from training smaller-scale models from scratch across a wide range of hyperparameters to analyzing publicly available, pretrained models to test our diagnostic framework in a realistic setting. Our primary selection criterion was the availability of two official pretrained weight versions, typically labeled "v1" and "v2", within the PyTorch model repository.

This v1/v2 setup provides a unique opportunity for a controlled comparison. By design, the v2 weights offer higher in-distribution (ID) accuracy on ImageNet, often due to improved training recipes, data augmentation (e.g., AutoAugment), or regularization (e.g., label smoothing). This allows us to directly test our central hypothesis: whether ID geometric markers can identify cases where higher ID accuracy masks a hidden vulnerability, leading to poorer out-of-distribution (OOD) transferability.

Our final set of 20 architectures is highly diverse, spanning multiple design generations and principles. In addition to deeper variants of models used in our control studies (\textbf{ResNet-50/101/152}, \textbf{MobileNetV2/V3}, \textbf{EfficientNet-B1}), our selection also includes:

\begin{itemize}[leftmargin=*]
\item \textbf{RegNet}~\citep{radosavovic2020designing}: A family of networks (e.g., RegNetY-400MF, RegNetX-32GF) whose structure is discovered by optimizing a data-driven design space, resulting in well-performing models.

\item \textbf{ResNeXt}~\citep{xie2017aggregatedresidualtransformationsdeep}: An evolution of ResNet that introduces a cardinality dimension, increasing model capacity by aggregating a set of parallel transformations.

\item \textbf{Wide ResNet}~\citep{zagoruyko2016wide}: A variant of ResNet that is wider but shallower, demonstrating that width can be a more effective dimension for improving performance than depth.
\end{itemize}

\subsection{Computing resources}
All experiments were conducted on NVIDIA H100 (80GB) or A100 (80GB) GPUs, paired with a 128-core Rome CPU and 1~TB of RAM. Training each model for 200 epochs required approximately 1--3~hours, depending on the architecture and optimizer. Unless otherwise specified, all experiments were run on a single GPU worker. These specifications, together with the full training configurations described in earlier subsections, are provided to facilitate reproducibility.

\newpage

\section{Details on ID Measures}
\label{app:measures}

In this section, we define the performance, statistical, and geometric measures used in our analysis. These are computed on the feature representations extracted from models using the ID training dataset, unless stated otherwise. Our goal is to identify which properties of a model's ID representations can serve as reliable indicators of its out-of-distribution (OOD) generalization capability.

The measures are grouped into three categories: \textbf{performance} measures that quantify classification accuracy, \textbf{statistical} measures that summarize low-order distributional properties of features, and \textbf{geometric} measures that characterize the structure of class-specific feature manifolds. A key distinction is that while statistical metrics typically operate on pooled features, our primary geometric measures are computed on object manifolds -- the per-class point clouds in representation space. This allows them to directly capture properties relevant to classification, such as manifold size, shape, and correlation structure in the representational space.

We first describe how feature representations are extracted and then define each measure in detail.

\subsection{Representation Extraction}
All representational measures are computed on feature vectors extracted from the penultimate layer of each network -- the final layer before the classification head. This layer captures high-level, task-specialized features that are not yet collapsed into class logits. For convolutional networks, the feature vector is obtained via global average pooling. The exact layers used for each architecture are listed in Table~\ref{tab:layer_selection}.

\begin{table}[ht]
  \caption{Exact layer names used for extracting feature representations.}
  \label{tab:layer_selection}
  \vspace{1mm}
  \centering
  \begin{tabular}{ll}
    \toprule
    Architecture & Layer name in PyTorch module \\
    \midrule
    VGG13 & \texttt{features.34} \\
    VGG16 & \texttt{features.43} \\
    VGG19 & \texttt{features.52} \\
    ResNet & \texttt{avgpool} \\
    DenseNet121 & \texttt{avg\_pool2d} \\
    MobileNet & \texttt{avg\_pool2d} \\
    EfficientNetB0 & \texttt{adaptive\_avg\_pool2d} \\
    RegNet & \texttt{avgpool} \\
    ResNeXt & \texttt{avgpool} \\
    Wide ResNet & \texttt{avgpool} \\
    \bottomrule
  \end{tabular}
\end{table}

Given an ID dataset $\mathcal{D}_{\text{ID}}$ and a trained network $f_\theta$, let $\mathbf{z}_i \in \mathbb{R}^N$ denote the $N$-dimensional feature vector for the $i$-th input sample $\mathbf{x}_i$ in $\mathcal{D}_{\text{ID}}$, extracted from the layer listed in Table~\ref{tab:layer_selection}. All statistical and geometric measures described in the following subsections are computed from the collection $\{\mathbf{z}_i\}_{i=1}^M$ of such feature vectors, where $M$ is the total number of samples in $\mathcal{D}_{\text{ID}}$.  

For measures that require class-specific statistics (e.g., within-class covariance, manifold radius), we further partition $\{\mathbf{z}_i\}$ by ground-truth label into $\{\mathbf{z}_i^{\mu}\}_{i=1}^{M^\mu}$ for each class $\mu \in \{1, \dots, P\}$, where $M^\mu$ is the number of samples in class $\mu$.

\subsection{Statistical metrics}\label{app:stat_measures}
We compute a set of statistical descriptors from the ID feature representations to quantify basic structural properties of the learned embedding space. All metrics are computed from the collection of penultimate-layer feature vectors $\{\mathbf{z}_i\}_{i=1}^M$ extracted from the ID dataset (see Table~\ref{tab:layer_selection}).  

\paragraph{Activation sparsity.}  
The activation sparsity measures the proportion of non-zero entries across all feature vectors,  
\[
\text{sparsity} = \frac{1}{MN} \sum_{i=1}^M \sum_{j=1}^N \mathbf{1}\!\left(|z_{ij}| > \varepsilon\right),
\]
where $N$ is the feature dimension and $\varepsilon = 10^{-6}$ is a small threshold to account for numerical noise. Higher sparsity indicates more silent units on average across the dataset.

\paragraph{Covariance magnitude.}  
We compute the empirical covariance matrix $\mathbf{\Sigma} \in \mathbb{R}^{N \times N}$ over features and take the mean absolute value of its off-diagonal entries,  
\[
\text{mean\_covariance} = \frac{2}{N(N-1)} \sum_{j<k} |\Sigma_{jk}|,
\]
which reflects the average degree of linear correlation between distinct feature dimensions.

\paragraph{Pairwise distance.}  
We compute the mean Euclidean distance between all pairs of feature vectors,  
\[
\text{mean\_distance} = \frac{2}{M(M-1)} \sum_{i<j} \|\mathbf{z}_i - \mathbf{z}_j\|_2,
\]
providing a coarse measure of spread in the representation space.

\paragraph{Pairwise angle.}  
After $\ell_2$-normalizing each feature vector, we compute cosine similarities and convert them to angles in radians via $\theta_{ij} = \arccos(\cos\_sim_{ij})$. The mean pairwise angle reflects the typical directional separation between features.

All statistical metrics are computed on the raw feature vectors without centering unless required by the measure (e.g., covariance).

\subsection{Geometric measures: participation ratio and GLUE-based task-relevant metrics}
\label{app:geom_measures}

Unlike the statistical measures described above, our geometric analysis operates on \textit{object manifolds}—point clouds in feature space containing activations from the same class. This distinction is important: geometric metrics explicitly quantify per-class representational structure, whereas most statistical metrics aggregate across the entire dataset without regard to class boundaries.

\paragraph{Participation ratio (PR).}
As a conventional baseline for manifold dimensionality, we compute the \textit{participation ratio} (PR) of the penultimate-layer features for each class. Let $\{\bz^\mu_i\}_{i=1}^{M^\mu}$ denote the $M^\mu$ feature vectors for the $\mu$-th class, and $\lambda_i^\mu$ be the eigenvalues of their covariance matrix. The PR of this class is defined as
\begin{equation}
D_{\textsf{PR}}^\mu = \frac{\left(\sum_{i} \lambda^\mu_i\right)^2}{\sum_{i} (\lambda_i^\mu)^2},
\end{equation}
which measures the effective number of principal components with substantial variance. In all figures we present the average of PR over all classes, i.e., $\frac{1}{P}\sum M^\mu D_{\textsf{PR}}^\mu$. While PR is widely used, it is \textit{task-agnostic} and does not incorporate information about class separability.

\paragraph{Neural Collapse measure (NC1).}
In addition to per-class geometric descriptors, we also include a global Neural Collapse--inspired
measure that captures the degree of \emph{zero-collapse} between within-class and between-class
structure~\citep{papyan2020prevalence,haruncontrolling}.  
Let $\Sigma_W \in \mathbb{R}^{N \times N}$ denote the pooled within-class covariance
and $\Sigma_B \in \mathbb{R}^{N \times N}$ the between-class covariance of the penultimate-layer
features (see \autoref{app:stat_measures} for definitions), and let $P$ be the number of classes.
We first form a truncated pseudo-inverse of $\Sigma_B$ by eigendecomposition.  
Write
\[
\Sigma_B \;=\; U \Lambda U^\top,
\quad
\Lambda = \mathrm{diag}(\lambda_1, \dots, \lambda_N),
\quad
\lambda_1 \ge \cdots \ge \lambda_N \ge 0,
\]
and let $\lambda_{\max} = \lambda_1$.
We retain only eigen-directions with sufficiently large eigenvalues,
\[
\mathcal{I} \;=\; \{\, i \;:\; \lambda_i \ge \tau \lambda_{\max} \,\},
\]
with a small threshold $\tau$ (we use $\tau = 10^{-3}$ in all experiments), and define the truncated
pseudo-inverse
\[
\Sigma_B^{\dagger}
\;=\;
\sum_{i \in \mathcal{I}} \lambda_i^{-1} \, \bu_i \bu_i^\top,
\]
where $\bu_i$ denotes the $i$-th column of $U$.
The NC1 (zero-collapse) score is then
\[
\mathrm{NC1}
\;=\;
\frac{1}{P} \, \mathrm{tr}\!\big( \Sigma_W \Sigma_B^{\dagger} \big).
\]
Smaller values of NC1 indicate stronger collapse of within-class variability relative to the
between-class structure.  
We treat NC1 as a geometric marker and compare it with the participation ratio, numerical rank,
and the GLUE-based task-relevant measures in our prognostic analysis.

\paragraph{Tunnel Effect: numerical rank.}
Inspired by recent studies on the Tunnel Effect hypothesis~\citep{masarczyk2023tunnel,harun2024variables}, we also compute the \textit{numerical rank} of the feature representations.
For a given class $\mu$, let $\{ \bz_i^\mu \}_{i=1}^{M^\mu}$ denote its feature vectors and let
\[
\Sigma_\mu 
\,=\, 
\frac{1}{M^\mu}
\sum_{i=1}^{M^\mu}
(\bz_i^\mu - \bc_\mu)(\bz_i^\mu - \bc_\mu)^\top
\]
be the corresponding empirical covariance matrix,
where $\bc_\mu$ is the class-mean representation (defined above).
Let $\sigma_1^\mu \ge \sigma_2^\mu \ge \cdots$ denote the singular values of $\Sigma_\mu$.
Following prior work, the numerical rank of class $\mu$ is defined as
\[
\mathrm{Rank}_{\mathrm{num}}^\mu 
\,=\, 
\#\left\{ i : \sigma_i^\mu \,\ge\, \tau \,\sigma_1^\mu \right\},
\quad\text{with}\;\; \tau = 10^{-3}.
\]
The reported value is the average over all classes,
$\mathrm{Rank}_{\mathrm{num}} = \frac{1}{P}\sum_{\mu=1}^P \mathrm{Rank}_{\mathrm{num}}^\mu$.
Lower numerical rank indicates stronger compression of the class manifold.
Prior work has shown that layers exhibiting low rank often display degraded OOD linear-probe accuracy.
We include numerical rank as a baseline geometric marker for comparison against the task-relevant GLUE-based measures.

\subsubsection{Task-relevant geometric measures from GLUE}
To capture the aspects of representational geometry most relevant for classification, we employ the effective geometric measures introduced in the \textit{Geometry Linked to Untangling Efficiency} (GLUE) framework~\citep{chou2025glue}, grounded in manifold capacity theory~\citep{chou2025glue,chung2018classification}. The theory has found wide applications in both neuroscience~\citep{yao2023transformation,paraouty2023sensory,kuoch2024probing,hu2024representational} and machine learning~\citep{cohen2020separability,mamou2020emergence,stephenson2021geometry,kirsanov2025,chou2025featurelearninglazyrichdichotomy}.

Analogous to support vector machine (SVM) theory—where an analytical connection between the max-margin linear classifier and its support vectors is used to assess separability in the \textit{best-case sense}—GLUE establishes a similar analytical connection in an \textit{average-case sense}, as follows.  
Rather than analyzing the max-margin classifier directly in the original $N$-dimensional feature space $\mathbb{R}^N$, GLUE considers random projections to an $N'$-dimensional subspace and evaluates whether the representations remain linearly separable. Intuitively, if the data are highly separable in $\mathbb{R}^N$, they will, with high probability, remain separable even after projection to a much lower $N'$. Conversely, if the data are barely separable in $\mathbb{R}^N$, the probability of maintaining separability will rapidly drop to zero as $N'$ decreases.  

Formally, following the modeling and notation in GLUE, each object manifold is modeled as the convex hull of all representations corresponding to the $\mu$-th class:  
\[
\mathcal{M}^{\mu} = \mathrm{conv}\left(\{\mathbf{z}^{\mu}_i\}_{i=1}^M\right),
\]  
where $\{\mathbf{z}^{\mu}_i\}$ is the collection of $M$ feature vectors of the $\mu$-th class.  
A dichotomy vector $\mathbf{y} \in \{-1,1\}^P$ and a collection $\mathcal{Y} \subset \{-1,1\}^P$ are chosen by the analyst. Common choices are $\mathcal{Y}$ being the set of all 1-vs-rest dichotomies (e.g., $(1,-1,-1,\dots,-1)$, $(-1,1,-1,\dots,-1)$, \dots, $(-1,-1,-1,\dots,1)$) or $\mathcal{Y} = \{-1,1\}^P$.  

The key quantity in GLUE for measuring the degree of (linear) separability of manifolds is the \textit{critical dimension}, defined as the smallest $N'$ such that the probability of (linear) separability after projection to a random $N'$-dimensional subspace is at least $0.5$:  
\[
N_{\mathrm{crit}} := \min_{p(N') \geq 0.5} N',
\]  
where  
\[
p(N') := \Pr_{\Pi:\mathbb{R}^N \to \mathbb{R}^{N'}}\left[\exists\,\mathbf{w} \in \mathbb{R}^{N'} \ \text{s.t.} \ y^\mu \langle \mathbf{w}, \mathbf{x}^\mu \rangle \geq 0, \ \forall \mu, \ \mathbf{x}^\mu \in \mathcal{M}^\mu\right].
\]  
By scaling $N_{\mathrm{crit}}$ with the number of manifolds, we define the \textit{classification capacity} $\alpha := P / N_{\mathrm{crit}}$, which intuitively captures the maximal load a network can handle. Larger $\alpha$ corresponds to more separable manifolds in the average-case sense.  

GLUE theory relates $\alpha$ to manifold structure through:  
\begin{equation}\label{eq:capacity qp form}
\alpha = P \cdot \left(\Exp_{\substack{\mathbf{y} \sim \mathcal{Y} \\ \mathbf{t} \sim \mathcal{N}(0, I_N)}} \left[\max_{\lambda^\mu_i \geq 0 \ \forall \mu,i} \left(\frac{\langle \mathbf{t}, \sum_{\mu,i}y^\mu \lambda^\mu_i \mathbf{z}^\mu_i \rangle}{\left\|\sum_{\mu,i} y^\mu \lambda^\mu_i \mathbf{z}^\mu_i \right\|_2}\right)^2 \right] \right)^{-1}.
\end{equation}  
Equation~\ref{eq:capacity qp form} can be numerically estimated using a quadratic programming solver (see Algorithm 1 in~\citep{chou2025glue}).  

Observe that one can view the optimal solution $\lambda^\mu(\by,\bt)$ for the inner maximization problem as a function of $\by,\bt$. This naturally leads to the following definition of \textit{anchor point} for class $\mu$ as:  
\[
\mathbf{s}^\mu(\mathbf{y}, \mathbf{t}) := \frac{\sum_i \lambda^\mu_i(\mathbf{y}, \mathbf{t}) \mathbf{z}^\mu_i}{\sum_i \lambda^\mu_i(\mathbf{y}, \mathbf{t})},
\]  
and stacking them into a matrix $\mathbf{S} \in \mathbb{R}^{P \times N}$ and let $\bS_\by:=\diag(\by)\bS$, GLUE yields an equivalent form:  
\begin{equation}\label{eq:capacity anchor form}
\alpha = P \cdot \left(\Exp_{\substack{\mathbf{y} \sim \mathcal{Y} \\ \mathbf{t} \sim \mathcal{N}(0, I_N)}} \left[(\bS_\by \mathbf{t})^\top (\bS_\by \bS_\by^\top)^{\dagger} (\bS_\by \mathbf{t})\right] \right)^{-1},
\end{equation}  
where $\dagger$ denotes the pseudoinverse. This parallels SVM theory, where the margin is linked to a simple function on the support vectors.  

\textbf{Center–axis decomposition of anchor points.}  
For each $\mu \in [P]$, define the anchor center of the $\mu$-th manifold as:  
\[
\mathbf{s}^\mu_0 := \mathbb{E}_{\mathbf{y},\mathbf{t}}\left[ \mathbf{s}^\mu(\mathbf{y},\mathbf{t}) \right],
\]  
and for each $(\mathbf{y},\mathbf{t})$, define the axis component of the $\mu$-th anchor point as:  
\[
\mathbf{s}^\mu_1(\mathbf{y},\mathbf{t}) := \mathbf{s}^\mu(\mathbf{y},\mathbf{t}) - \mathbf{s}^\mu_0.
\]  
Similar to $\bS_\by$, we denote $\bS_{\by,0}, \bS_{\by,1}(\mathbf{y},\mathbf{t}) \in \mathbb{R}^{P \times N}$ as the matrices containing $y^\mu\mathbf{s}^\mu_0$ and $y^\mu\mathbf{s}^\mu_1(\mathbf{y},\mathbf{t})$ on their rows, respectively, i.e., $\bS_{\by,0}:=\diag(\by)\bS_0$ and $\bS_{\by,1}(\mathbf{y},\mathbf{t}):=\diag(\by)\bS_1(\by,\bt)$ where $\bS_0$ and $\bS_1(\by,\bt)$ have $\mathbf{s}^\mu_0$ and $\mathbf{s}^\mu_1(\mathbf{y},\mathbf{t})$ stacked on their rows.

With these, define:  
\begin{align*}
a(\mathbf{y},\mathbf{t}) &= (\bS_\by\mathbf{t})^\top(\bS_\by\bS_\by^\top)^{\dagger}(\bS_\by\mathbf{t}), \\
b(\mathbf{y},\mathbf{t}) &= (\bS_{\by,1}(\mathbf{y},\mathbf{t})\mathbf{t})^\top \left(\bS_{\by,1}(\mathbf{y},\mathbf{t})\bS_{\by,1}(\mathbf{y},\mathbf{t})^\top\right)^{\dagger} (\bS_{\by,1}(\mathbf{y},\mathbf{t})\mathbf{t}), \\
c(\mathbf{y},\mathbf{t}) &= (\bS_{\by,1}(\mathbf{y},\mathbf{t})\mathbf{t})^\top \left(\bS_{\by,0}\bS_{\by,0}^\top + \bS_{\by,1}(\mathbf{y},\mathbf{t})\bS_{\by,1}(\mathbf{y},\mathbf{t})^\top\right)^{\dagger} (\bS_{\by,1}(\mathbf{y},\mathbf{t})\mathbf{t}).
\end{align*}  
Note that $\alpha = P / \mathbb{E}_{\mathbf{y},\mathbf{t}}[a(\mathbf{y},\mathbf{t})]$.  

\textbf{Effective geometric measures.}  
GLUE further decomposes $\alpha$ into three measures:  
\[
\alpha = \Psi_\eff \cdot \frac{1 + R_\eff^{-2}}{D_\eff},
\]  
where:  
\begin{itemize}
    \item \textbf{Effective dimension:}  
    \[
    D_\eff := \frac{1}{P} \, \mathbb{E}_{\mathbf{y},\mathbf{t}}[b(\mathbf{y},\mathbf{t})]
    \]  
    Intuitively, $D_\eff$ measures the intrinsic dimensionality of the manifolds while incorporating \textit{axis alignment} between them. Lower $D_\eff$ corresponds to more compact, better-aligned manifolds, improving linear separability.  

    \item \textbf{Effective radius:}  
    \[
    R_\eff := \sqrt{\frac{\mathbb{E}_{\mathbf{y},\mathbf{t}}[c(\mathbf{y},\mathbf{t})]}{\mathbb{E}_{\mathbf{y},\mathbf{t}}[b(\mathbf{y},\mathbf{t}) - c(\mathbf{y},\mathbf{t})]}}
    \]  
    Intuitively, $R_\eff$ quantifies the scale of manifold variation relative to its center, incorporating \textit{center alignment} between classes. Smaller $R_\eff$ reflects tighter clustering of features around class centers, reducing manifold overlap.  

    \item \textbf{Effective utility:}  
    \[
    \Psi_\eff := \frac{\mathbb{E}_{\mathbf{y},\mathbf{t}}[c(\mathbf{y},\mathbf{t})]}{\mathbb{E}_{\mathbf{y},\mathbf{t}}[a(\mathbf{y},\mathbf{t})]}
    \]  
    Intuitively, $\Psi_\eff$ measures the combined effect of \textit{signal-to-noise ratio} (SNR) on separability. Higher $\Psi_\eff$ corresponds to manifolds that are both low-dimensional and compact relative to inter-class distances.
\end{itemize}  

For further derivations, illustrations, and examples, see the supplementary materials of~\citep{chou2025glue}. In all our experiments, for each manifold we subsample to 50 points, conduct GLUE analysis on each manifold pair, and apply Gaussianization preprocessing~\citep{wakhloo2023linear} to ensure initial linear separability.



\[
\rho^c_{\mu,\nu} := |\langle\bs^\mu_0,\bs^\nu_0\rangle|
\]

\[
\rho^a_{\mu,\nu} := \Exp_{\by,\bt}[|\langle\bs^\mu_1(\by,\bt),\bs^\nu_1(\by,\bt)\rangle|]
\]

\[
\psi_{\mu,\nu} := \Exp_{\by,\bt}[|\langle\bs^\mu_0,\bs^\nu_1(\by,\bt)\rangle|]
\]

\paragraph{Implementation details.}
In all our experiments, we consider the following specific hyperparameter choice for GLUE analysis. We randomly

\paragraph{Intuitions for GLUE measures.}
The three task-relevant geometric measures—$D_\eff$, $R_\eff$, and $\Psi_\eff$—serve as markers that directly link geometric properties of object manifolds to classification efficiency. As we show in later sections, they are substantially more predictive of OOD performance than conventional measures. Here we summarize key properties, examples, and approximations of GLUE measures in~\autoref{tab:effective geometry} for intuition-building.

\begin{table}[h]
\centering
\footnotesize
\caption{Intuitions for GLUE measures.}
\label{tab:effective geometry}
\vspace{2mm}
\renewcommand{\arraystretch}{1.3} 
\begin{tabular}{|p{2cm}|p{3.3cm}|p{3.5cm}|p{3.3cm}|}
\hline
& \multicolumn{1}{c|}{$D_\eff \geq 0$} &  \multicolumn{1}{c|}{$R_\eff \geq 0$} & \multicolumn{1}{c|}{$\Psi_\eff \in [0,1]$} \\ \hline
Geometric \ \ \ \ \ \ \ \ intuition & Quantify the task-relevant dimensionality of object manifolds. & Quantify the task-relevant spread within each manifold relative to their centers. & Quantify the amount of excessive compression of untangling manifolds . \\ \hline
Effect on linear separability & More separable when $D_\eff$ is small. & More separable when $R_\eff$ is small. & More separable when $\Psi_\eff$ is large. \\ \hline
Example &  $D_\eff$ equals the dimension of uncor.~random spheres & $R_\eff$ equals the radius of uncor.~random spheres & Collapsing manifolds to points yields $\Psi_\eff\to0$. \\ \hline
Formula for uncorrelated random spheres\footnotemark & \vspace{0.1mm} $\frac{1}{P}\sum_\mu\Exp\left[\left(\frac{\langle\bs^\mu_1(\bt),\bt\rangle}{\|\bs^\mu_1(\bt)\|_2}\right)^2\right]$ & \vspace{0.1mm} $\frac{1}{P}\sum_\mu\sqrt{\Exp\left[\left(\frac{\|\bs^\mu_1(\bt)\|_2}{\|\bs^\mu_0\|_2}\right)^2\right]}$ \vspace{0.1mm} & \vspace{0.1mm} $\frac{1}{P}\sum_\mu\Exp\left[\left(\frac{\langle\bs^\mu_1(\bt),\bt\rangle}{\langle\bs^\mu(\bt),\bt\rangle}\right)^2\right]$ \\ \hline
Interaction with correlations among manifolds & If within-manifold variations align along similar directions, $D_\eff$ decreases. & If manifold centers move farther apart, $R_\eff$ decreases. & If within-manifold variations reduce without improving separability, $\Psi_\eff$ decreases. \\ \hline
Interpretation in feature learning\footnotemark & Low $D_\eff$ indicates a smaller set of feature modes in use. &  Low $R_\eff$ indicates more similar feature usage across examples within a class. & Low $\Psi_\eff$ indicates inefficient compression of within-class variability. \\ \hline
\end{tabular}
\end{table}

\footnotetext[13]{For the $\mu$-th manifold, define its anchor center as $\bs^\mu_0:=\Exp_\bt[\bs^\mu(\bt)]$ and the axis-part of the anchor point as $\bs^\mu_1(\bt):=\bs^\mu(t)-\bs^\mu_0$. Intuitively, $\bs^\mu_0$ is the mean representation for the $\mu$-class, and $\bs^\mu_1(\bt)$ corresponds to the within-class variation/spread. $\langle\cdot,\cdot\rangle$ denotes inner product and $\|\cdot\|_2$ denotes $\ell_2$ norm. Formulas for uncorrelated random spheres provide a useful mental picture: $D_\eff$ resembles the \textit{Gaussian width}, equal to the sphere’s dimension~\citep{vershynin2018high}; $R_\eff$ reflects the ratio of within-manifold variation to mean response; and $\Psi_\eff$ corresponds to the fraction of error (i.e., inner product with $\bt$) attributable to within-manifold variation. 
}

\footnotetext[14]{We follow a top-down view of feature learning~\citep{chou2025featurelearninglazyrichdichotomy}, where \textit{features} are understood functionally through their consequences for computation (e.g., enabling linear separability) rather than as specific interpretable axes or neurons. This perspective emphasizes how representational geometry changes with feature usage without requiring explicit identification of the features themselves. Moreover, by thinking of a direction in the representation space as a feature (linear representation hypothesis~\citep{park2024linear}), the effective geometric measures offer interpretation in feature learning as listed in the table.}

\clearpage

\section{Additional Results for~\autoref{sec:prognostic}}
\label{app:additional_results}

\subsection{Implementation details}

During the initial exploration of how OOD performance varies across a wide range of final model states, we trained all architectures from scratch on CIFAR-10. We used two optimizers: SGD with a momentum of 0.9, and AdamW~\citep{loshchilov2018decoupled}. We ran training for 200 epochs with a cosine annealing learning rate schedule, which smoothly decays the learning rate to zero, stabilizing late-stage representation geometry.

For each architecture and optimizer pair, we performed a systematic $4 \times 4$ grid search over the initial learning rate ($\eta_0$) and weight decay ($\lambda$). The specific values for each grid, which were tailored to each architecture family based on empirical best practices, are detailed in Table~\ref{tab:grid_sgd} and Table~\ref{tab:grid_adamw}. This diverse grid was designed to produce models in various training regimes, from under- to over-regularized, allowing us to find cases where ID performance is stable while OOD performance varies --- a key aspect of our analysis.

\begin{table}[ht]
\small
  \caption{Hyperparameter grid for SGD optimizer.}
  \label{tab:grid_sgd}
  \vspace{1mm}
  \centering
  \begin{tabular}{lll}
    \toprule
    Architecture & Initial learning rate list & Weight decay list \\
    \midrule
    VGG (13/16/19) & [0.01000, 0.00333, 0.00111, 0.00037] & [0.0010000, 0.0003333, 0.0001111, 0.0000370] \\
    ResNet (18/34/50) & [1.00000, 0.50000, 0.25000, 0.12500] & [0.0002000, 0.0001000, 0.0000500, 0.0000250] \\
    DenseNet121 & [0.05000, 0.01667, 0.00556, 0.00185] & [0.0005000, 0.0001667, 0.0000556, 0.0000185] \\
    MobileNet & [0.20000, 0.06667, 0.02222, 0.00741] & [0.0001000, 0.0000333, 0.0000111, 0.0000037] \\
    EfficientNetB0 & [0.20000, 0.06667, 0.02222, 0.00741] & [0.0001000, 0.0000333, 0.0000111, 0.0000037] \\
    \bottomrule
  \end{tabular}
\end{table}

\begin{table}[ht]
\small
  \caption{Hyperparameter grid for AdamW optimizer.}
  \label{tab:grid_adamw}
  \vspace{1mm}
  \centering
  \begin{tabular}{lll}
    \toprule
    Architecture & Initial learning rate list & Weight decay list \\
    \midrule
    VGG (13/16/19) & [0.02000, 0.00500, 0.00125, 0.00031] & [0.0100000, 0.0033333, 0.0011111, 0] \\
    ResNet (18/34/50) & [0.10000, 0.02500, 0.00625, 0.00156] & [0.0100000, 0.0050000, 0.0025000, 0] \\
    DenseNet121 & [0.05000, 0.02500, 0.01250, 0.00625] & [0.0100000, 0.0033333, 0.0011111, 0] \\
    MobileNet & [0.02000, 0.00500, 0.00125, 0.04000] & [0.0100000, 0.0033333, 0.0011111, 0] \\
    EfficientNetB0 & [0.10000, 0.05000, 0.01000, 0.00100] & [0.0010000, 0.0003333, 0.0001111, 0] \\
    \bottomrule
  \end{tabular}
\end{table}

\subsection{Details for~\autoref{fig:main across more params}}
In this section, we present supplementary figures that provide a more detailed view of the main findings stated in~\autoref{fig:main across more params} from~\autoref{sec:prognostic}.~\autoref{tab:additional_results_structure} provides a list of content for this subsection.

\paragraph{Quantification of Relationships.}  

We quantify the relationship between ID measures and OOD performance by computing the Pearson correlation coefficient ($r$) and its associated $p$-value via ordinary least-squares linear regression between the measure values and OOD accuracies. For all figures with heatmaps, we annotate each $r$-value with significance asterisks based on its $p$-value: $p \le 0.0001$ ($^{****}$), $p \le 0.001$ ($^{***}$), $p \le 0.01$ ($^{**}$), and $p \le 0.05$ ($^{*}$).

\begin{table}[ht]
    \caption{Organization of figures in Appendix~\ref{app:additional_results}.  
    }
    \label{tab:additional_results_structure}
    \vspace{1mm}
    \centering
    \begin{tabular}{lllll}
        \toprule
         Figure label & Model set & Optimizer & ID split & OOD dataset \\
        \midrule
         \autoref{fig:5dnn_sgd_train}        & Five DNNs                  & SGD    & Train & CIFAR-100 \\
         \autoref{fig:5dnn_sgd_test}         & Five DNNs                  & SGD    & Test  & CIFAR-100 \\
         \autoref{fig:5dnn_adamw_train}      & Five DNNs                  & AdamW  & Train & CIFAR-100 \\
         \autoref{fig:5dnn_adamw_test}       & Five DNNs                  & AdamW  & Test  & CIFAR-100 \\
         \autoref{fig:resnet_vgg_sgd_train}  & Three ResNets + Three VGGs & SGD    & Train & CIFAR-100 \\
         \autoref{fig:resnet_vgg_sgd_test}   & Three ResNets + Three VGGs & SGD    & Test  & CIFAR-100 \\
         \autoref{fig:imagenet_2dnn_sgd_train}& ResNet18 + VGG19                 & SGD    & Train & ImageNet subset \\
         \autoref{fig:imagenet_2dnn_sgd_test}& ResNet18 + VGG19                  & SGD    & Test  & ImageNet subset \\
        \bottomrule
    \end{tabular}
\end{table}

\paragraph{Data Splits for Measures.}
The terms ``Test'' and ``Train'' in the figure labels indicate whether the representational measures were computed on the ID test set or the ID training set, respectively.


\begin{figure}[h!]
    \centering
    \includegraphics[width=0.85\linewidth]{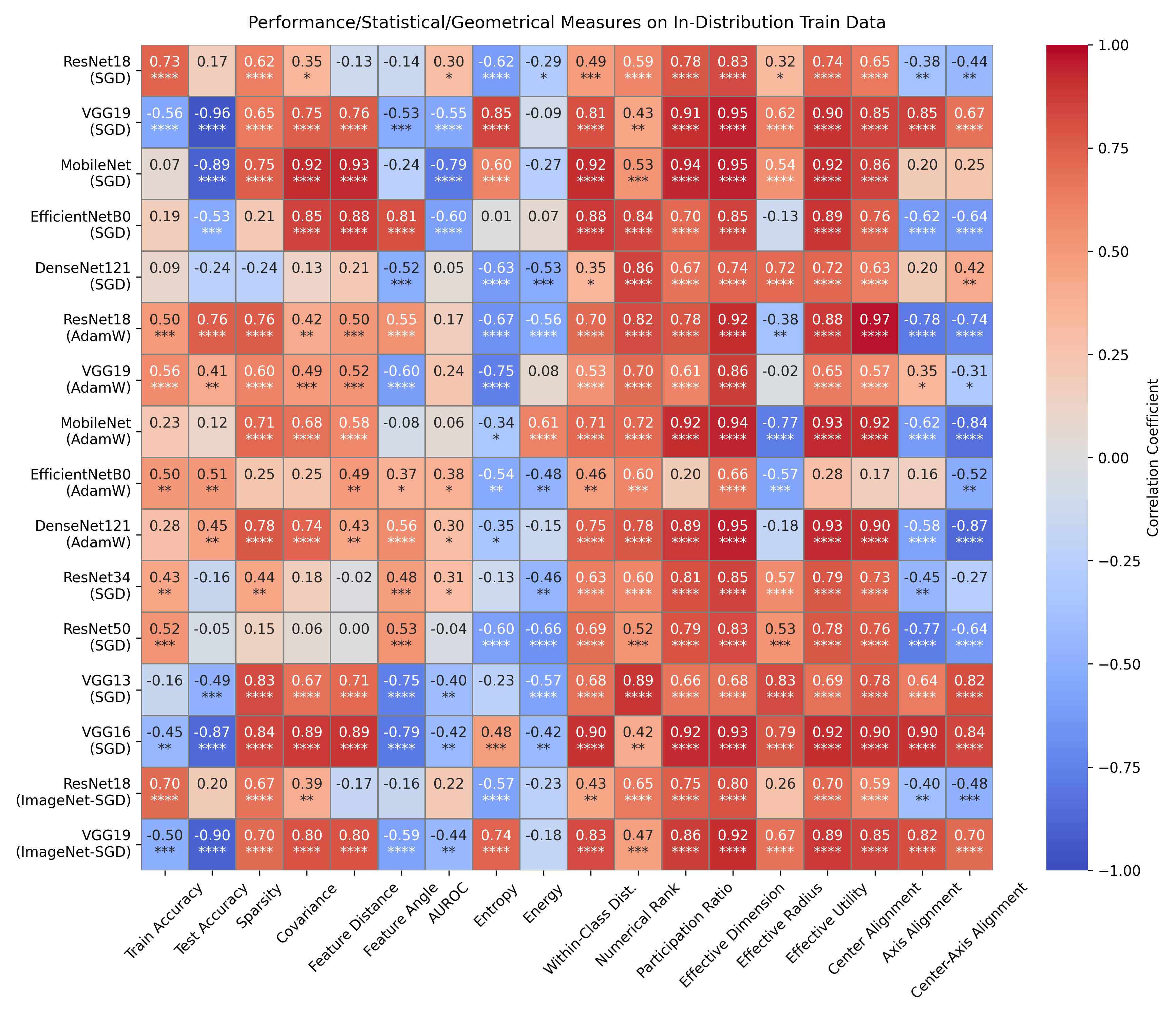}
    \vspace{-3mm}
    \caption{All results, measures computed on the ID \textit{train} set.}
    \label{fig:table_train}
\end{figure}
\vspace{-6mm}


\begin{figure}[h!]
    \centering
    \includegraphics[width=0.95\linewidth]{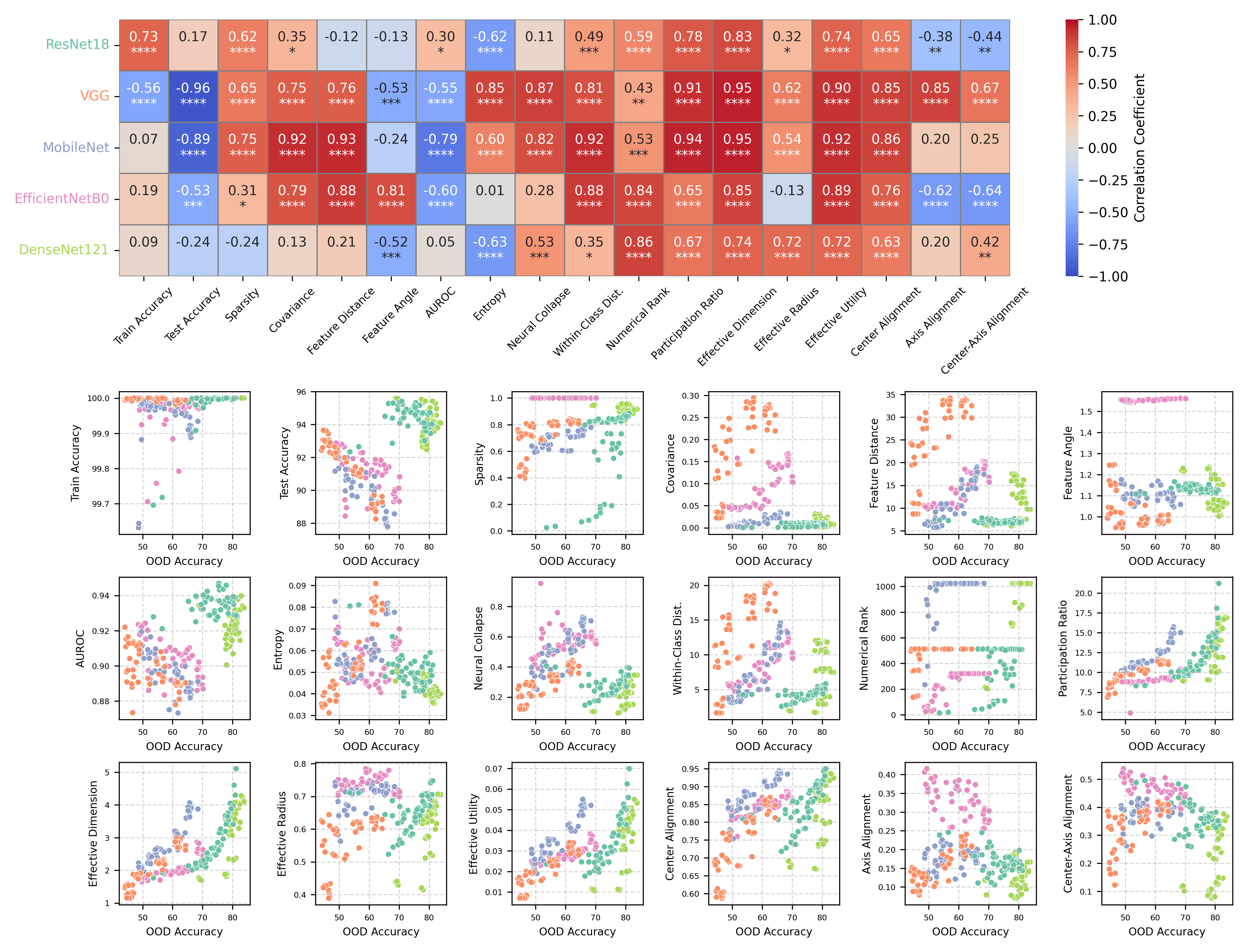}
    \caption{Five DNN architectures, trained with SGD, measures computed on the ID \textit{train} set.}
    \label{fig:5dnn_sgd_train}
\end{figure}

\begin{figure}[h!]
    \centering
    \includegraphics[width=0.95\linewidth]{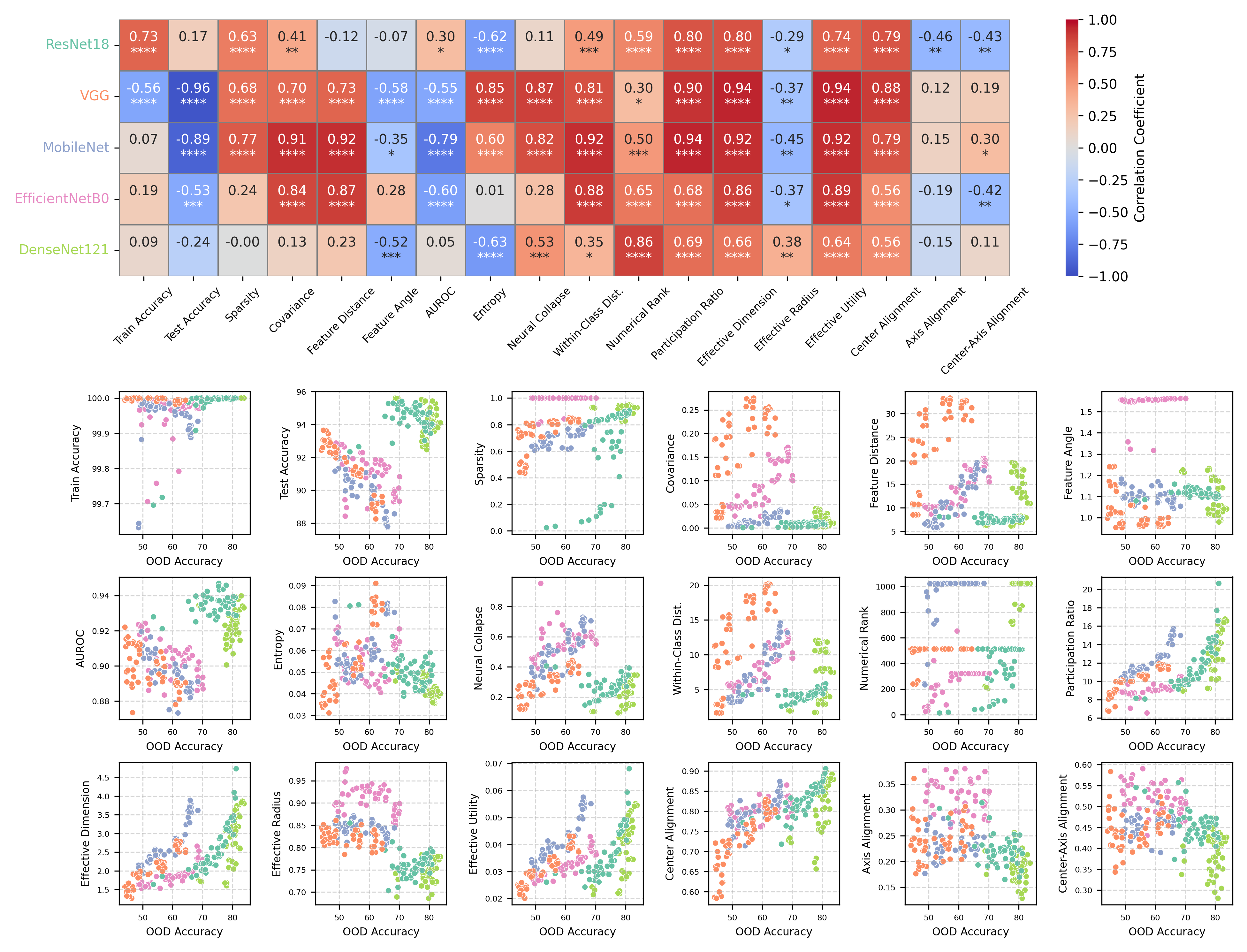}
    \caption{Five DNN architectures, trained with SGD, measures computed on the ID \textit{test} set.}
    \label{fig:5dnn_sgd_test}
\end{figure}

\begin{figure}[h!]
    \centering
    \includegraphics[width=0.95\linewidth]{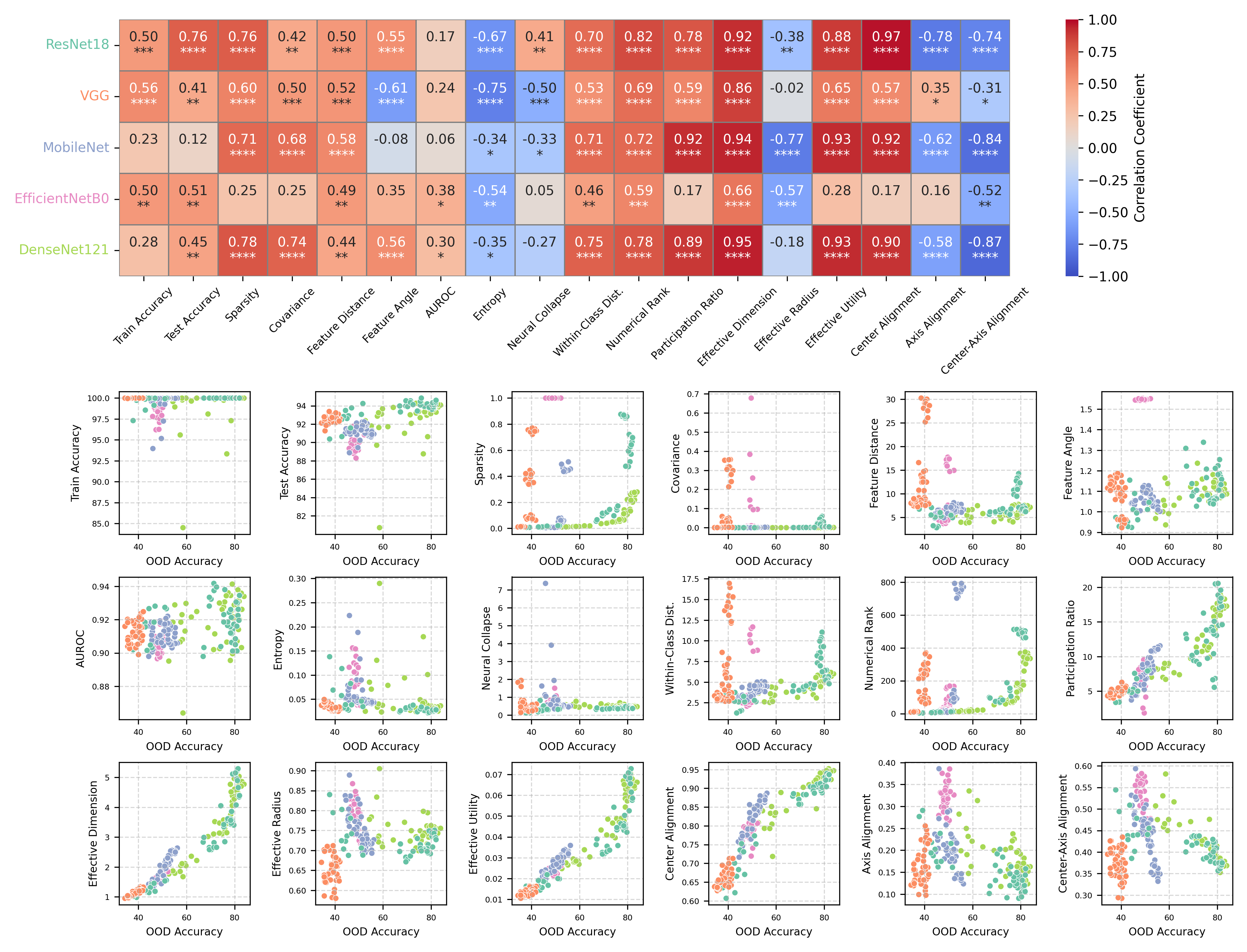}
    \caption{Five DNN architectures, trained with AdamW, measures computed on the ID \textit{train} set.}
    \label{fig:5dnn_adamw_train}
\end{figure}

\begin{figure}[h!]
    \centering
    \includegraphics[width=0.95\linewidth]{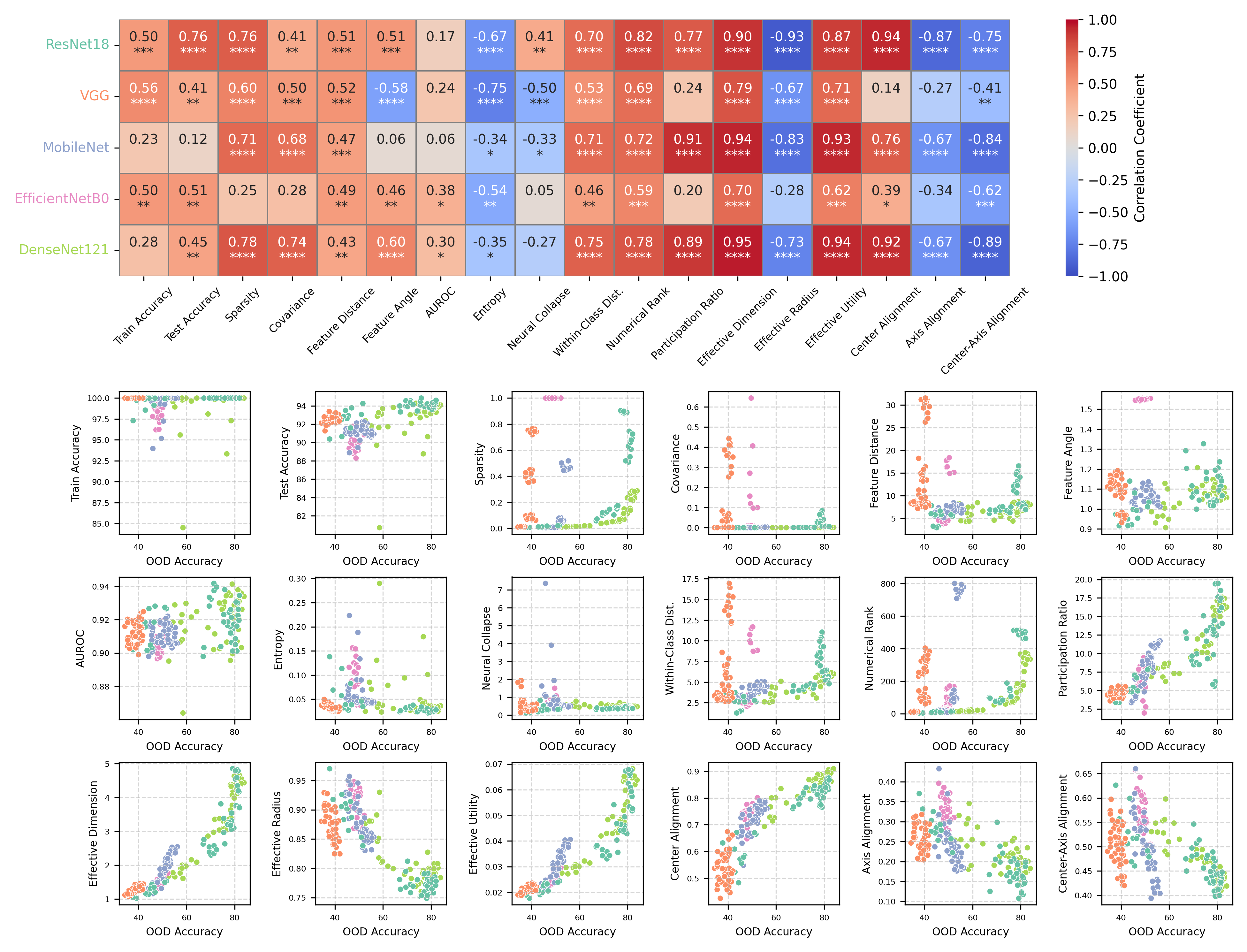}
    \caption{Five DNN architectures, trained with AdamW, measures computed on the ID \textit{test} set.}
    \label{fig:5dnn_adamw_test}
\end{figure}

\begin{figure}[h!]
    \centering
    \includegraphics[width=0.95\linewidth]{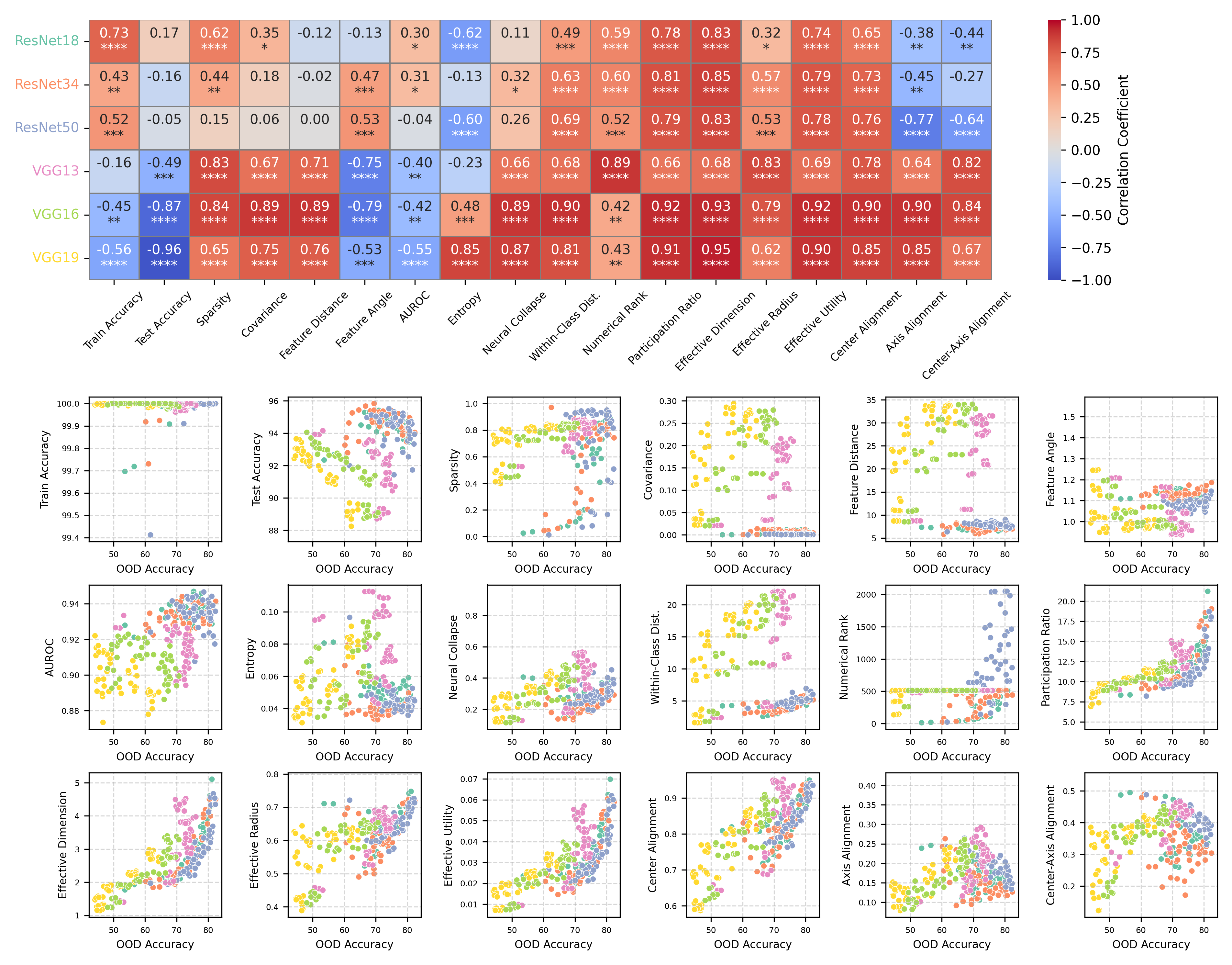}
    \caption{Three ResNet and three VGG architectures, trained with SGD, measures computed on the ID \textit{train} set.}
    \label{fig:resnet_vgg_sgd_train}
\end{figure}

\begin{figure}[h!]
    \centering
    \includegraphics[width=0.95\linewidth]{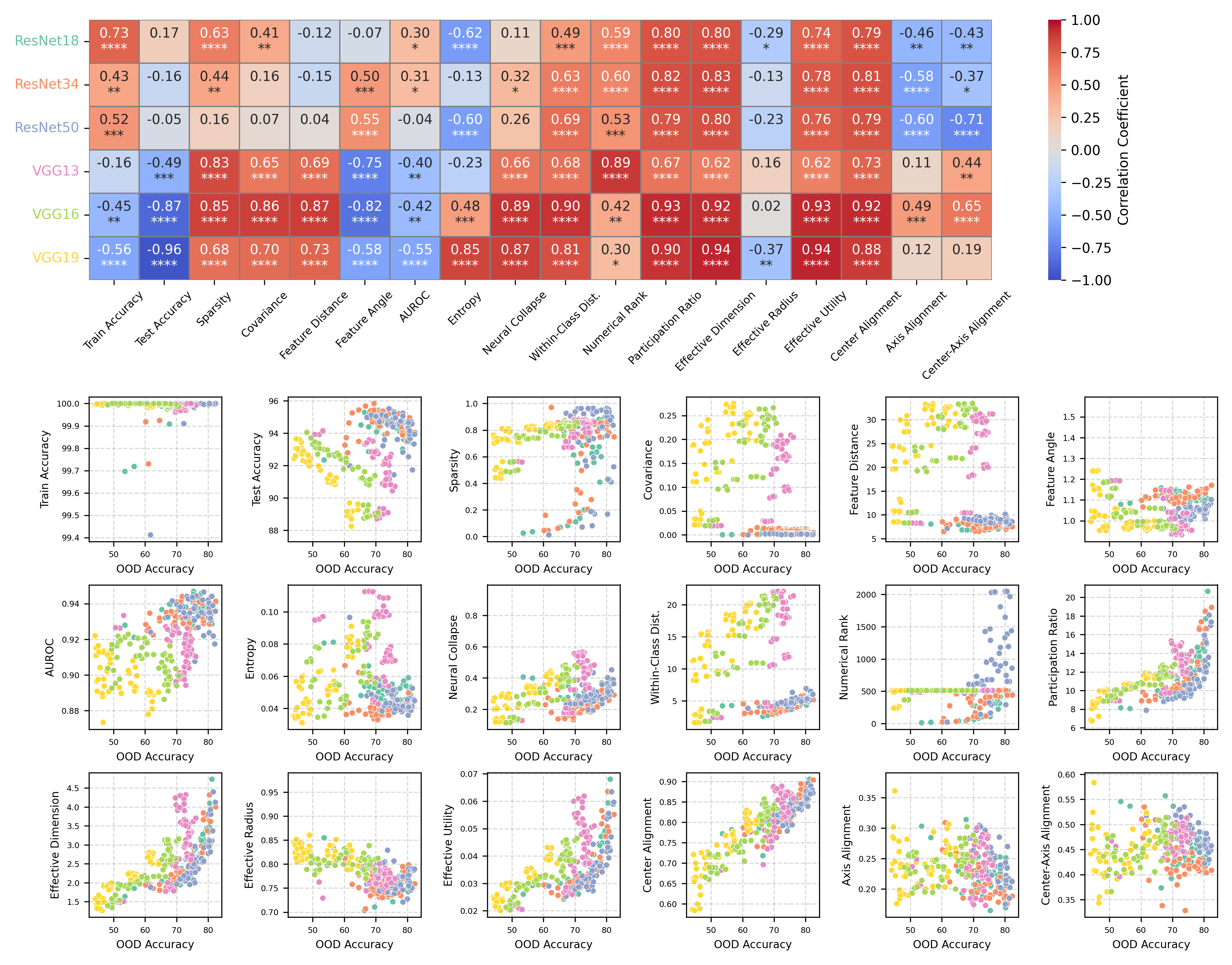}
    \caption{Three ResNet and three VGG architectures, trained with SGD, measures computed on the ID \textit{test} set.}
    \label{fig:resnet_vgg_sgd_test}
\end{figure}

\begin{figure}[h!]
    \centering
    \includegraphics[width=0.95\linewidth]{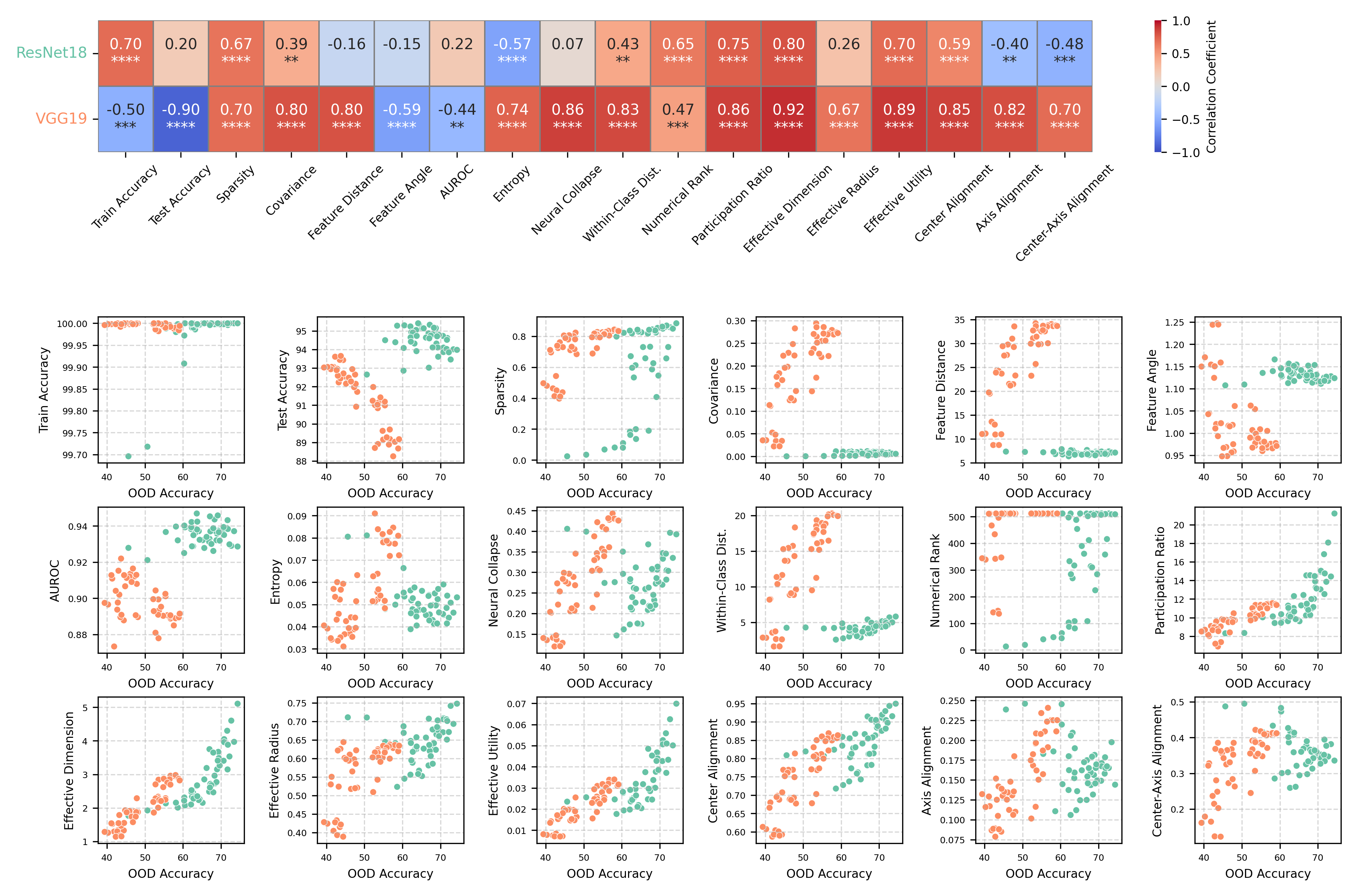}
    \caption{ResNet18 and VGG19, trained with SGD, evaluated on ImageNet subset OOD, measures computed on the ID \textit{train} set.}
    \label{fig:imagenet_2dnn_sgd_train}
\end{figure}

\begin{figure}[h!]
    \centering
    \includegraphics[width=0.95\linewidth]{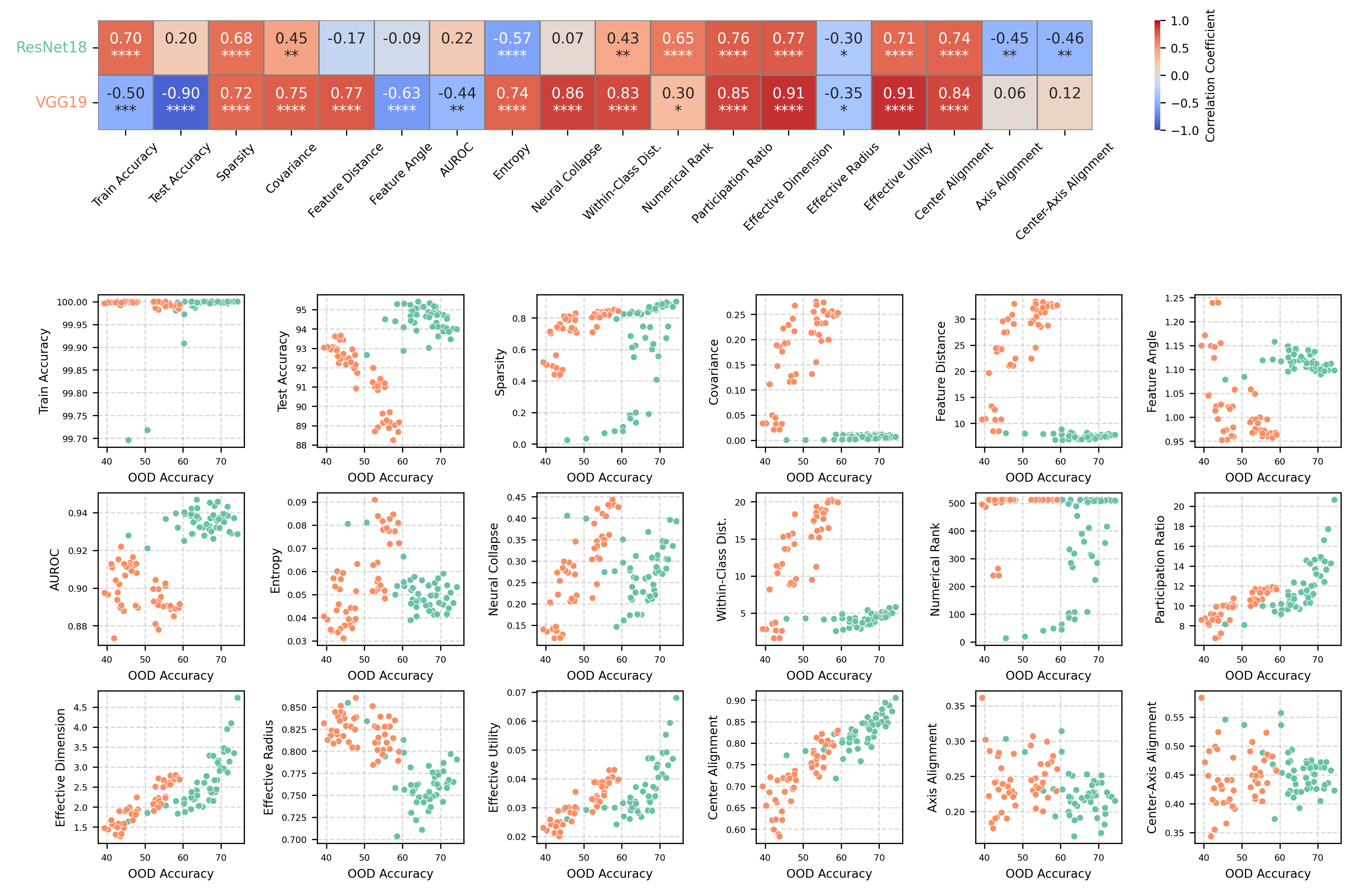}
    \caption{ResNet18 and VGG19, trained with SGD, evaluated on ImageNet subset OOD, measures computed on the ID \textit{test} set.}
    \label{fig:imagenet_2dnn_sgd_test}
\end{figure}


\clearpage
\subsection{Results on corrupted images as OOD data}~\label{app:corrupted data}
Here we provide results on 6 out of 19 corruption methods in CIFAR-10C.

\begin{figure}[h!]
    \centering
    \includegraphics[width=0.95\linewidth]{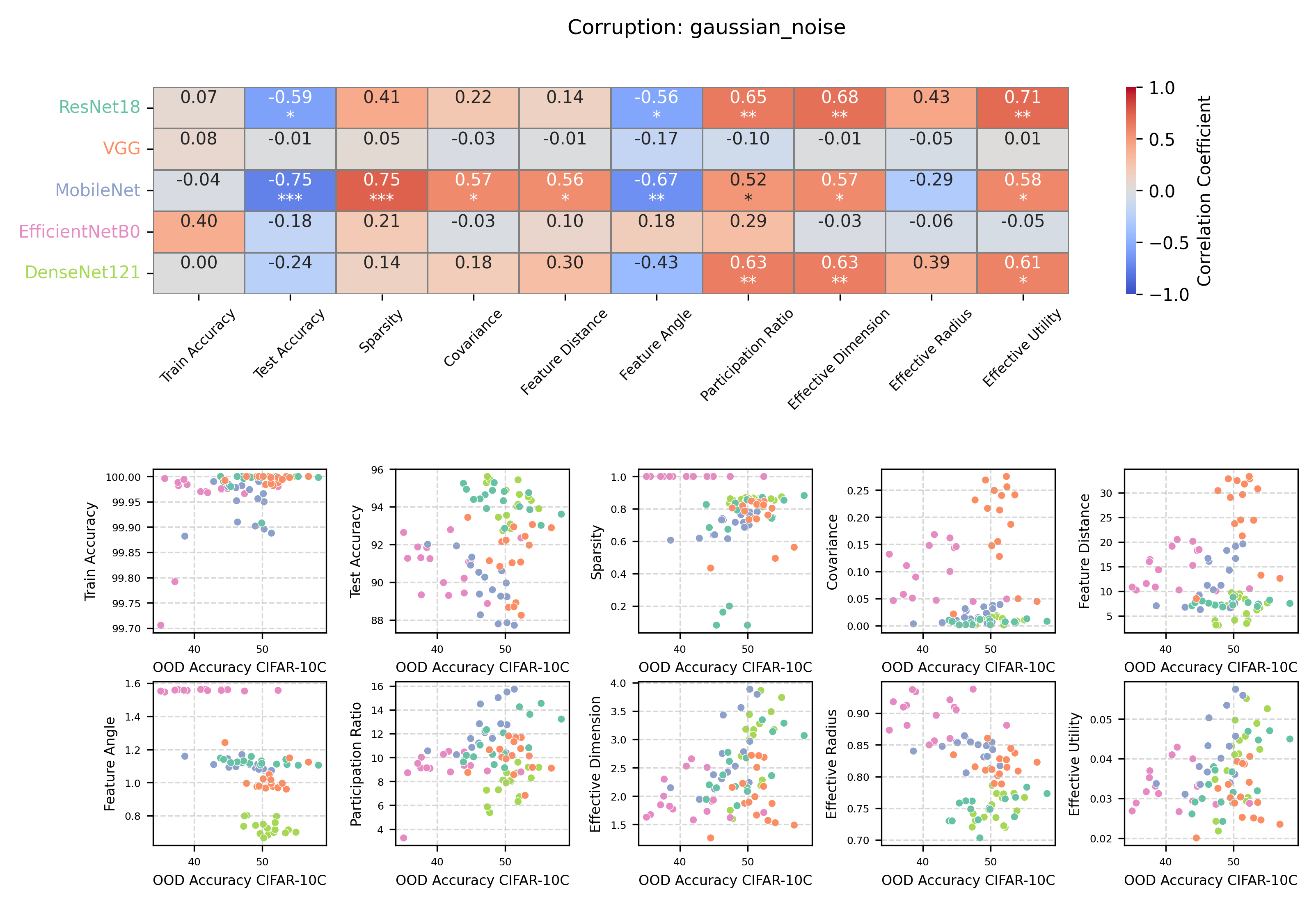}
    \caption{Corruption type: gaussian noise. Five DNN architectures, trained with SGD, measures computed on the ID \textit{test} set.}
    \label{fig:corruption gaussian noise}
\end{figure}

\begin{figure}[h!]
    \centering
    \includegraphics[width=0.95\linewidth]{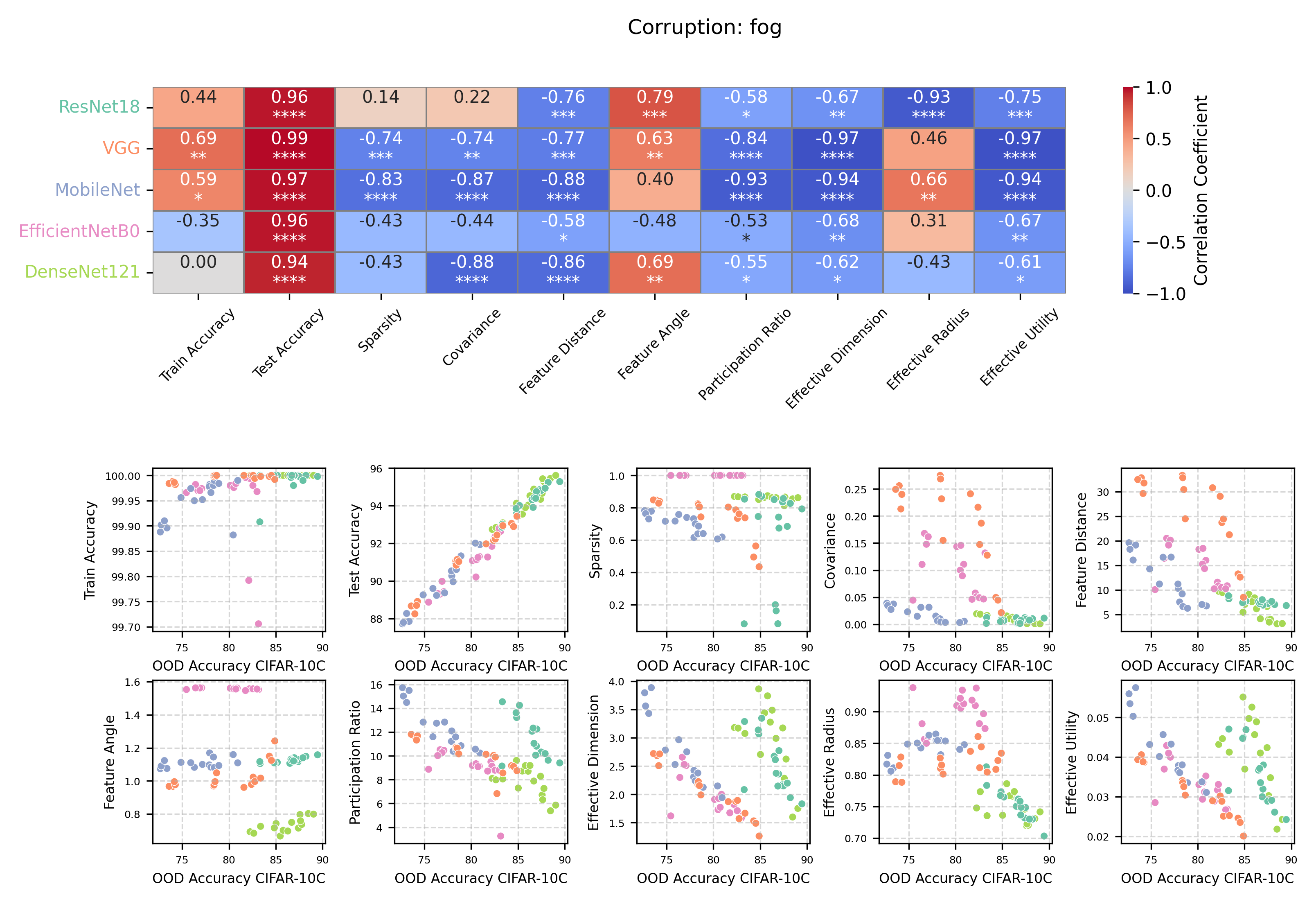}
    \caption{Corruption type: fog. Five DNN architectures, trained with SGD, measures computed on the ID \textit{test} set.}
    \label{fig:corruption fog}
\end{figure}

\begin{figure}[h!]
    \centering
    \includegraphics[width=0.95\linewidth]{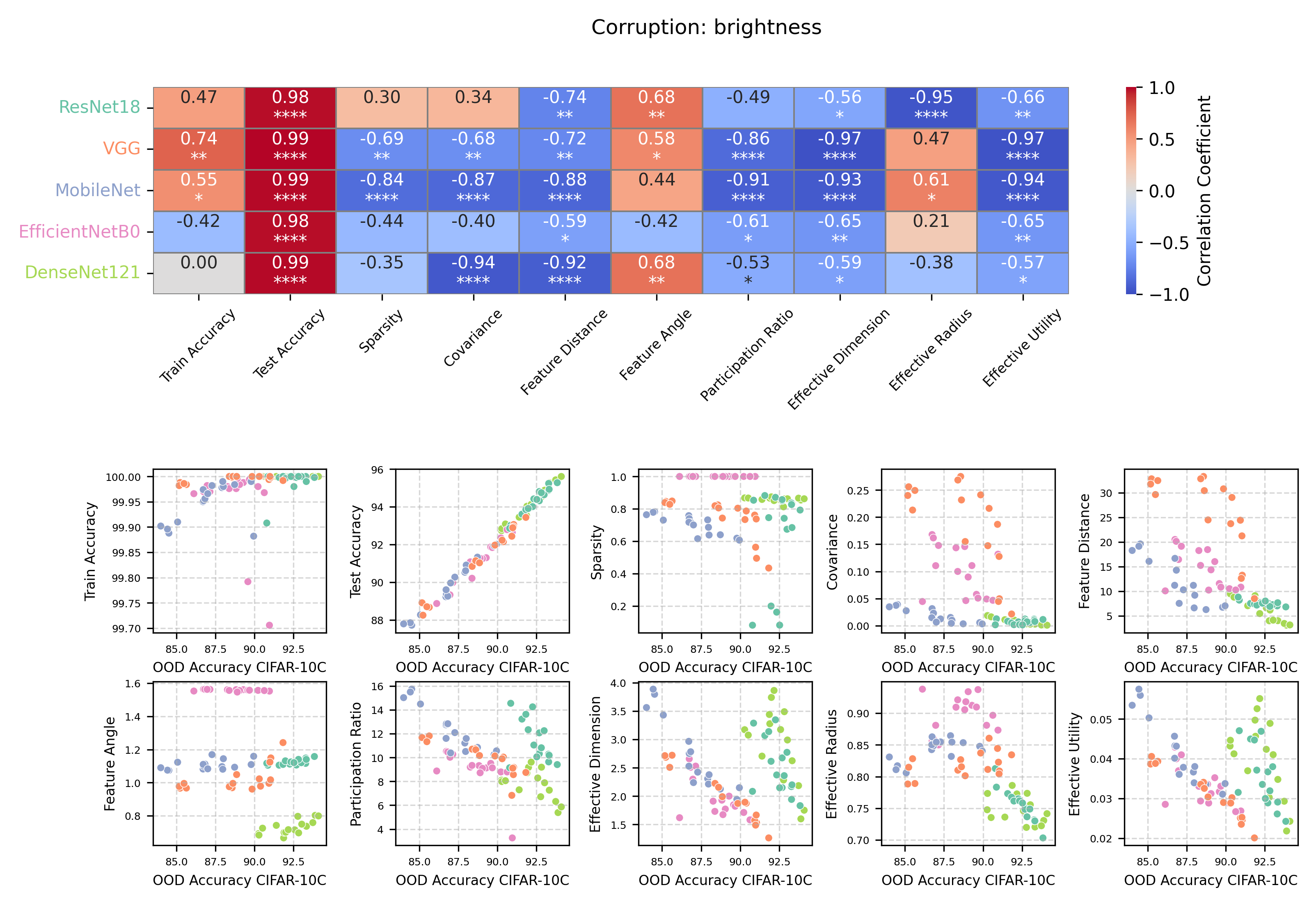}
    \caption{Corruption type: brightness. Five DNN architectures, trained with SGD, measures computed on the ID \textit{test} set.}
    \label{fig:corruption brightness}
\end{figure}

\begin{figure}[h!]
    \centering
    \includegraphics[width=0.95\linewidth]{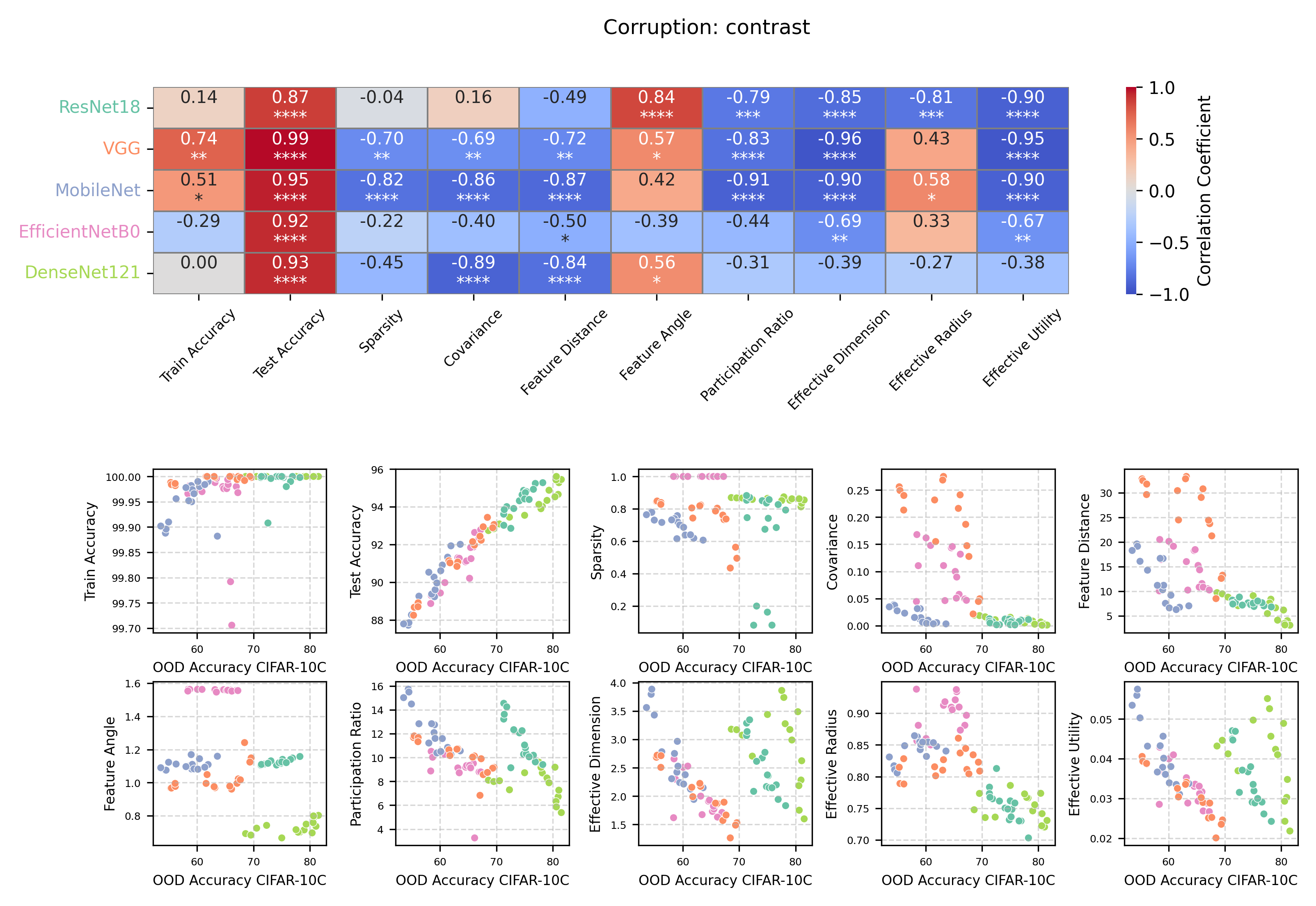}
    \caption{Corruption type: contrast. Five DNN architectures, trained with SGD, measures computed on the ID \textit{test} set.}
    \label{fig:corruption contrast}
\end{figure}

\begin{figure}[h!]
    \centering
    \includegraphics[width=0.95\linewidth]{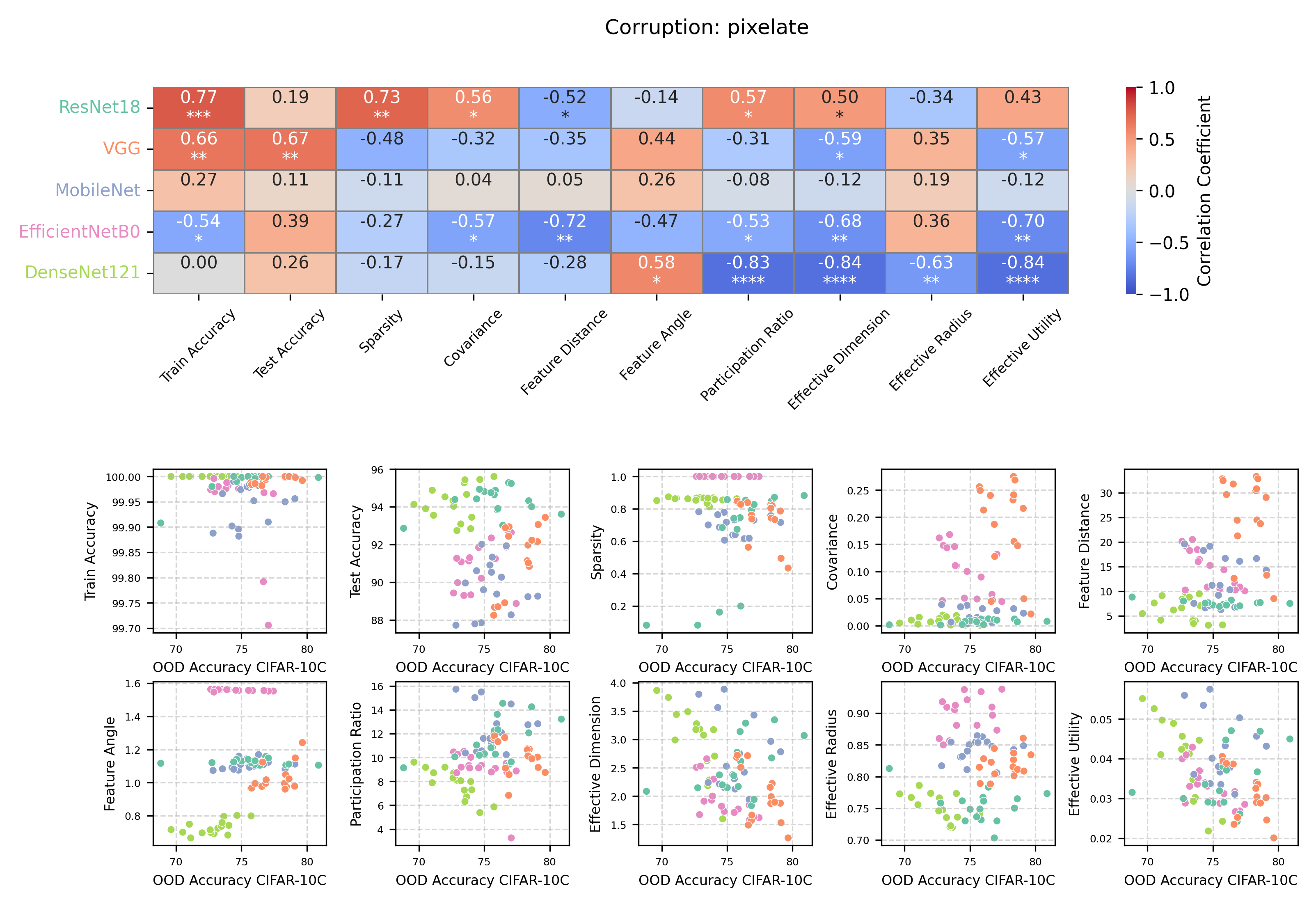}
    \caption{Corruption type: pixelate. Five DNN architectures, trained with SGD, measures computed on the ID \textit{test} set.}
    \label{fig:corruption pixelate}
\end{figure}

\begin{figure}[h!]
    \centering
    \includegraphics[width=0.95\linewidth]{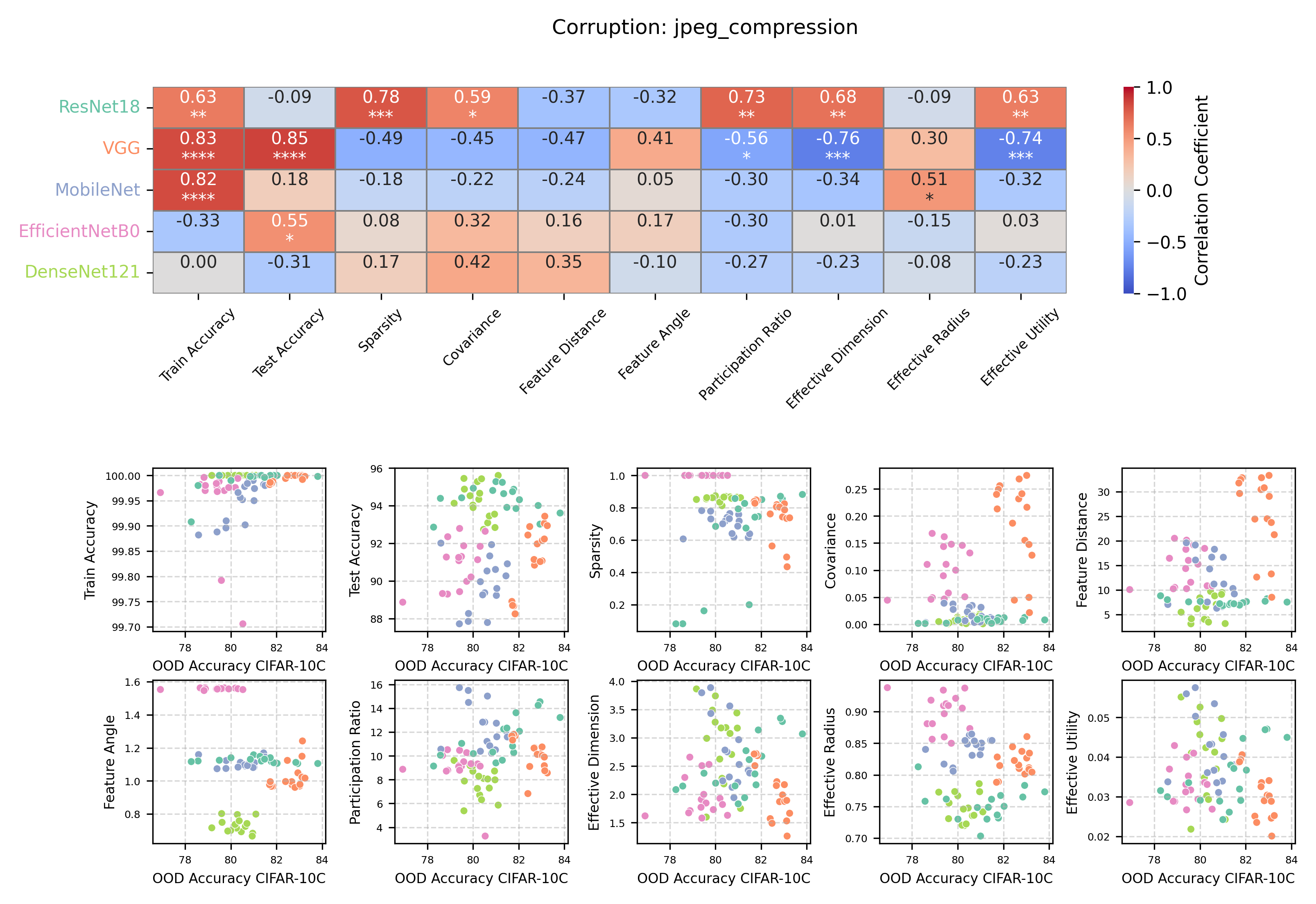}
    \caption{Corruption type: jpeg compression. Five DNN architectures, trained with SGD, measures computed on the ID \textit{test} set.}
    \label{fig:corruption jpeg compression}
\end{figure}

\clearpage
\section{Details on the Applications to Pretrained models}\label{app:applications}
Here we provide implementation details and statistical procedures underlying the pretrained model analysis in~\autoref{sec:application}. This section accompanies the full results reported in \autoref{fig:pretrained other} and Figures \ref{fig:finetuning_flowers102}, \ref{fig:finetuning_stanfordcars}.

\subsection{Model Selection and Weights}
We evaluated 20 pretrained architectures available through the PyTorch model zoo, spanning families such as RegNet, MobileNet, ResNet/ResNeXt, WideResNet, EfficientNet, and Vision Transformer (ViT). For each architecture, we included both the \texttt{v1} and \texttt{v2} weight releases. The two weight sets differ in training recipes and regularization schemes, though exact details are not always disclosed, making them a heterogeneous and realistic testbed. By design, the \texttt{v2} models typically achieve higher ImageNet top-1 accuracy, while \texttt{v1} weights often exhibit higher manifold dimensionality.

We remark that for the ViT models, we treat \texttt{IMAGENET1K\_SWAG\_LINEAR\_V1} as \texttt{v1} and \texttt{IMAGENET1K\_SWAG\_E2E\_V1} as \texttt{v2}.

\subsection{Representation Extraction}
For each model, we extracted feature representations from the penultimate layer (see Table~\ref{tab:layer_selection} for exact layer names). Input images were preprocessed by resizing to $224 \times 224$ pixels, converted to tensors, and normalized with standard ImageNet statistics. For GLUE analysis, we subsampled 2 classes and for each class we subsampled 50 feature vectors, applied Gaussianization preprocessing, and computed effective geometric measures ($D_{\text{eff}}, R_{\text{eff}}, \Psi_{\text{eff}}$) as described in Appendix~\ref{app:measures}. We repeated the above random subsampling for 100 times.

\subsection{OOD Evaluation via Linear Probing}
To evaluate the OOD generalization of the frozen feature extractor, we attached a linear classifier to the penultimate feature representation of each pretrained model (see Table~\ref{tab:layer_selection} for layer details). Crucially, the pretrained backbone weights remained frozen throughout this process; only the parameters of the new classifier were trained.
For each OOD dataset, we train linear classifiers on the penultimate feature vectors for $50$ epochs using the Adam optimizer with an initial learning rate of $0.1$ and a cross-entropy loss function. In all the results, we report the average linear probe accuracy over 3 repetitions on different random seeds.

\subsection{Prognostic prediction}\label{sec:pretrained prognstic pred}
For each model, after measuring the $(D_\eff,\Psi_\eff)$ of $\texttt{v1}$ and $\texttt{v2}$ respectively. We use the following criteria to make a prognostic prediction: if the $D_\eff(x)-D_\eff(y)$ is greater than the sum of the standard error of estimating $D_\eff(x)$ and $D_\eff(y)$, plus $\Psi_\eff(x)-\Psi_\eff(y)$ is greater than the sum of the standard error of estimating $\Psi_\eff(x)$ and $\Psi_\eff(y)$, then we predict $x$ is going to have better OOD performance than $y$; otherwise we make no verdict (here $x,y\in\{\texttt{v1},\texttt{v2}\}$).

Recall that in~\autoref{sec:application} we applied our prognostic method to 20 ImageNet-pretrained models across 9 OOD datasets and achieved a prediction accuracy of 73.02\% (compared to 37.22\% when using ID test accuracy as the marker).
Here, we systematically evaluate other markers that showed reasonable performance in~\autoref{sec:prognostic results}. Specifically, we consider $D_\eff$ and $\Psi_\eff$ as before, along with the Neural Collapse metric, numerical rank, average within-class distance, and participation ratio (definitions in~\autoref{app:measures}).

The prediction procedure follows the same criterion described earlier: for each marker, we compare the two weight versions ($\texttt{v1}$ vs.~$\texttt{v2}$) and issue a prediction only when the gap between their marker values exceeds the sum of the standard errors of estimation. We evaluate both individual markers and pairwise combinations.

\begin{figure}[h]
    \centering
    \includegraphics[width=\linewidth]{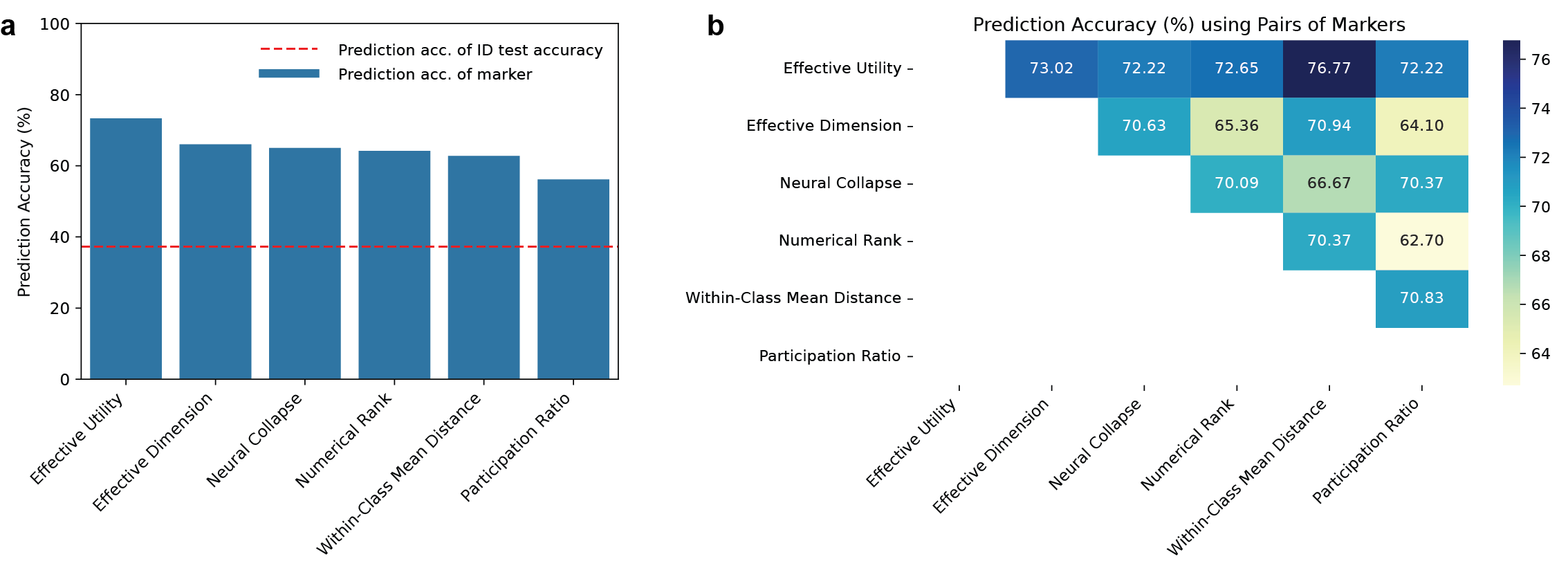}
    \caption{Prediction accuracy of OOD performance using different markers (or marker combinations). \textbf{a}, Using single marker. \textbf{b}, Using a pair of markers.}
    \label{fig:pretrained other}
\end{figure}

\begin{figure}[h!]
    \centering
    \includegraphics[width=0.9\linewidth]{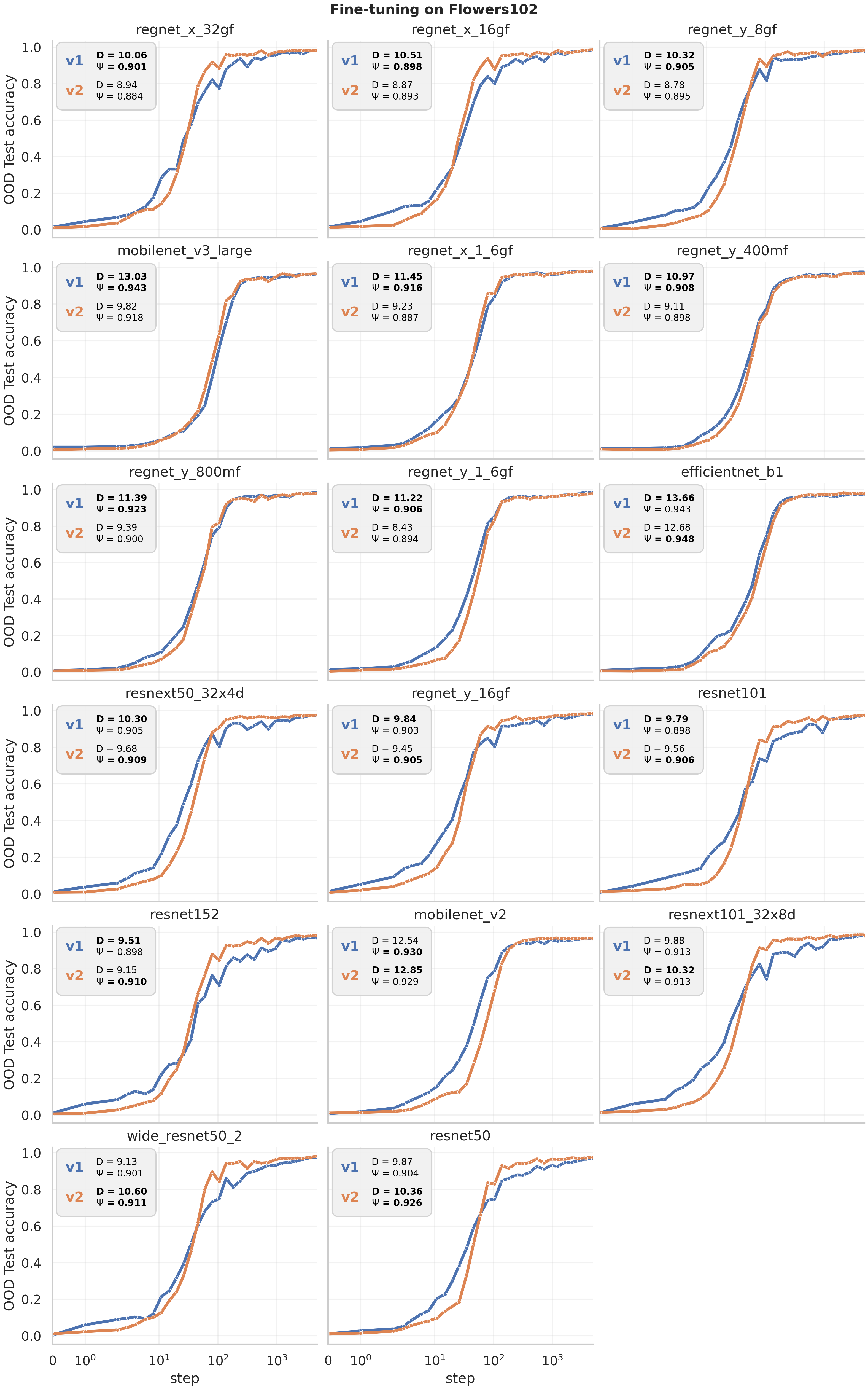}
    \caption{Fine-tuning dynamics of ImageNet-pretrained networks on \textbf{Flowers102} dataset from v1 and v2 weights. Insets show ID measures at initialization}
    \label{fig:finetuning_flowers102}
\end{figure}

\begin{figure}[h!]
    \centering
    \includegraphics[width=0.9\linewidth]{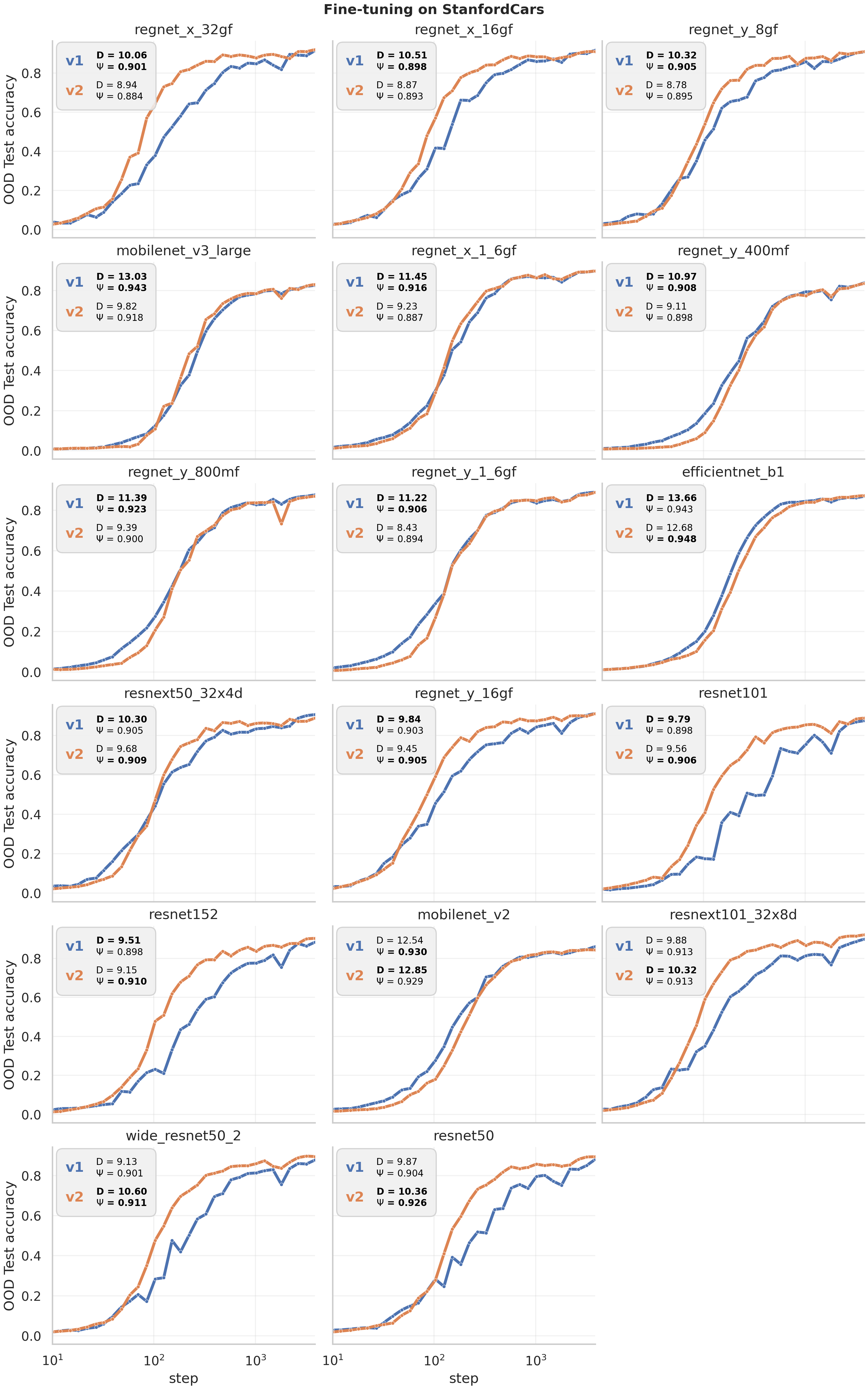}
    \caption{Fine-tuning dynamics of ImageNet-pretrained networks on \textbf{StanfordCars} dataset from v1 and v2 weights. Insets show ID measures at initialization}
    \label{fig:finetuning_stanfordcars}
\end{figure}

The results, summarized in~\autoref{fig:pretrained other}, show that all these markers substantially outperform ID test accuracy as prognostic indicators of OOD transfer performance.

\paragraph{Remark on alternative markers and future directions.}
As shown in Fig.~\ref{fig:pretrained other}, several alternative markers—or combinations of markers—also achieve strong prognostic performance, and in some cases perform comparably to or slightly better than the specific pair $(D_\eff,\Psi_\eff)$ used in the main analysis. This is fully consistent with the broader message of our work: a wide range of manifold-geometry-based quantities, both within and outside the GLUE family, contain significant predictive signal for OOD transfer performance. A deeper understanding of why different markers succeed on different subsets of architectures and how these markers may complement one another is an exciting direction for future investigation.

It is important to emphasize that the goal of the present experiment is not to identify a single “optimal” marker, but rather to demonstrate that geometric markers offer a substantial improvement over the conventional practice of using ID test accuracy as a predictor of OOD performance. Indeed, across all markers and marker-pairs we evaluated, the resulting prediction accuracies (ranging from 62\% to 76\%) consistently exceed that of ID test accuracy (37.22\%) by a factor of approximately two. This reinforces the central conclusion that geometry-based diagnostics provide a robust and broadly effective alternative for prognostic prediction in transfer learning.

\subsection{Full model fine-tuning protocol}
As a complementary evaluation, we also performed end-to-end fine-tuning. Models were initialized with either the v1 or v2 pretrained weights, and a new task-specific classifier head was randomly initialized. Unlike the linear probe, all model parameters (both in the backbone and the new classifier) were updated during training.

To simulate a realistic application scenario, we fine-tuned the models on the complete official training splits of Flowers102 (6,149 images) and Stanford Cars (8,144 images). Training was conducted for 50 epochs with a batch size of 64. We used the AdamW optimizer~\citep{loshchilov2018decoupled} with a weight decay of $10^{-6}$ and a cosine annealing learning rate scheduler with an initial learning rate of $3 \times 10^{-4}$.

To monitor the learning dynamics, we evaluated the model's performance on the validation set at 40 checkpoints, spaced logarithmically throughout the training process.

\end{document}